# Commonsense Reasoning in Computer Vision: Foundations, Recent Advancements, and Future Directions

BAHAR UDDIN MAHMUD*, Lander University, USA
SUMIT BARUA, Western Michigan University, USA
GUAN YUE HONG, Western Michigan University, USA
AJAY GUPTA, Western Michigan University, USA
HEXU LIU, Western Michigan University, USA

Commonsense reasoning in computer vision encompasses integrating visual data and contextual knowledge, crucial for enhancing AI's understanding of everyday scenarios. This understanding not only improves machine learning models but also enhances their ability to interact meaningfully with humans and the environment. Unlike CNN-based conventional vision models, which are designed to identify objects within a specific image, incorporating commonsense knowledge enables models to interpret scenes in a more holistic manner, thereby improving their spatial ability to reason about relationships among objects and actions. This integration not only enhances object recognition but also facilitates a deeper understanding of the contextual factors, ultimately leading to more precise predictions and interactions in real-world applications. This paper presents a comprehensive survey of recent developments that integrate commonsense knowledge into computer vision tasks. We systematically review approaches based on knowledge graphs, scene graphs, neuro-symbolic models, and commonsense-augmented transformers. We also outline current limitations related to dataset bias, knowledge incompleteness, and integration challenges. Finally, we highlight prospective research trajectories in cross-modal reasoning, scalable commonsense knowledge injection, and neuro-symbolic hybrid architectures to develop truly intelligent visual systems.

## 1 Introduction

Traditional computer vision models are designed to detect objects in an image by identifying and classifying parts of the image that were learned during model training. Those models are trained with supervised learning on large datasets containing labeled images, enabling them to learn pixel-level patterns and utilize loss functions to enhance their predictions.

Although traditional vision systems, such as Convolutional Neural Networks (CNNs), establish a strong foundation and have achieved remarkable success in object detection and classification, they fall short in

*Corresponding Author.

Authors' Contact Information: Bahar Uddin Mahmud, ORCID: 0000-0000-0000-000X, bmahmud@lander.edu, Lander University, Greenwood, SC, USA; Sumit Barua, ORCID: 0000-0000-0000-000X, sumit.barua@wmich.edu, Western Michigan University, Kalamazoo, MI, USA; Guan Yue Hong, ORCID: 0000-0000-0000-000X, guanyue.hong@wmich.edu, Western Michigan University, Kalamazoo, MI, USA; Ajay Gupta, ORCID: 0000-0000-0000-000X, ajay.gupta@wmich.edu, Western Michigan University, Kalamazoo, MI, USA; Hexu Liu, ORCID: 0000-0000-0000-000X, hexu.liu@wmich.edu, Western Michigan University, Kalamazoo, MI, USA.

understanding the underlying semantics of the scenes they analyze. CNN models can classify patterns by detecting edges, textures, shapes, and repeated visual patterns, but these models cannot understand the meaning of the objects they detect. Furthermore, multiple studies like Zellers et al. (2019) (Zellers et al. 2019) have highlighted the inconsistencies in the performance of CNN-based models on tasks that require inference beyond superficial visual cues, showing that models trained only with visual encoders perform poorly on the VCR benchmark, which involves answering complex questions about scenes and providing justifications for those answers. For example, a traditional CNN model can detect a cat in an image as shown in Figure 1, but the model does not understand whether the detected object is an animal, why it looks scared, or what its next move is. Similarly, in autonomous driving scenarios, a perception system might accurately detect a pedestrian at a crosswalk but fail to infer the pedestrian's intention to cross based on body language, gaze direction, or contextual cues like traffic light status. In medical imaging, a model may segment a tumor with high precision but cannot reason about its clinical significance relative to surrounding anatomical structures, patient history, or typical disease progression patterns. These limitations underscore the need for commonsense reasoning—the ability to integrate observable visual data with background knowledge about how the world typically works.

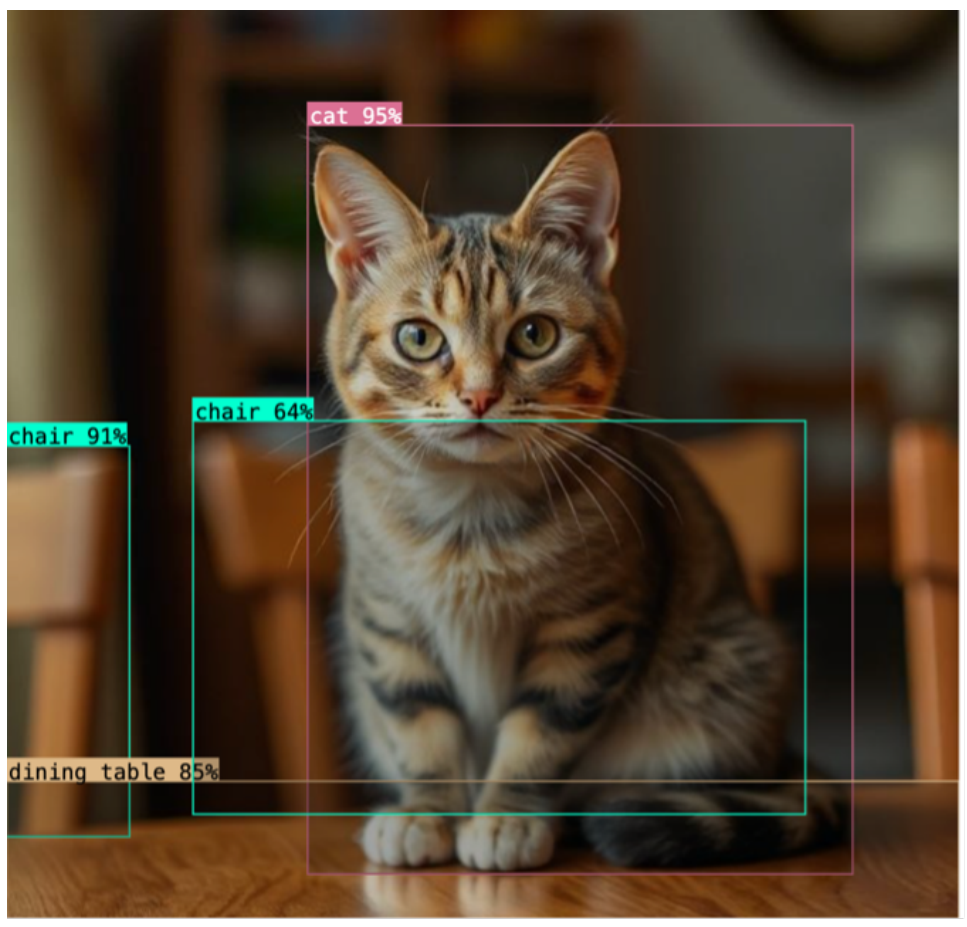

Fig. 1. CNN evaluation of a cat sitting on the table. The model can detect the cat with 95% accuracy, but is unable to tell us the meaning of its facial expression, which is essential to calculate its next move.

These limitations in traditional computer vision models motivated the development of more advanced models such as transformers, multimodal architectures, and foundation models. The integration of commonsense knowledge into vision models necessitates representing and reasoning over implicit information that is not directly observable. According to Sap et al. (2020) (Sap et al. 2019), commonsense reasoning in vision involves not only recognizing objects or actions in an image, but also predicting the likely causes and effects of events in the image that require knowledge of the physical and social world. Commonsense reasoning in AI can be mathematically formalized as the integration of observed evidence $X$ and background knowledge $K$ to predict outcomes $Y$, as expressed by:

$$P(Y|X,K) = P(Y|X) \times P(Y|K), \tag{1}$$

where $P(Y|X)$ captures the likelihood of the predicted outcome based on the observed data and $P(Y|K)$ represents the prior informed by external knowledge. This formulation emphasizes the importance of integrating direct perception with commonsense priors to achieve robust and generalizable inferences.

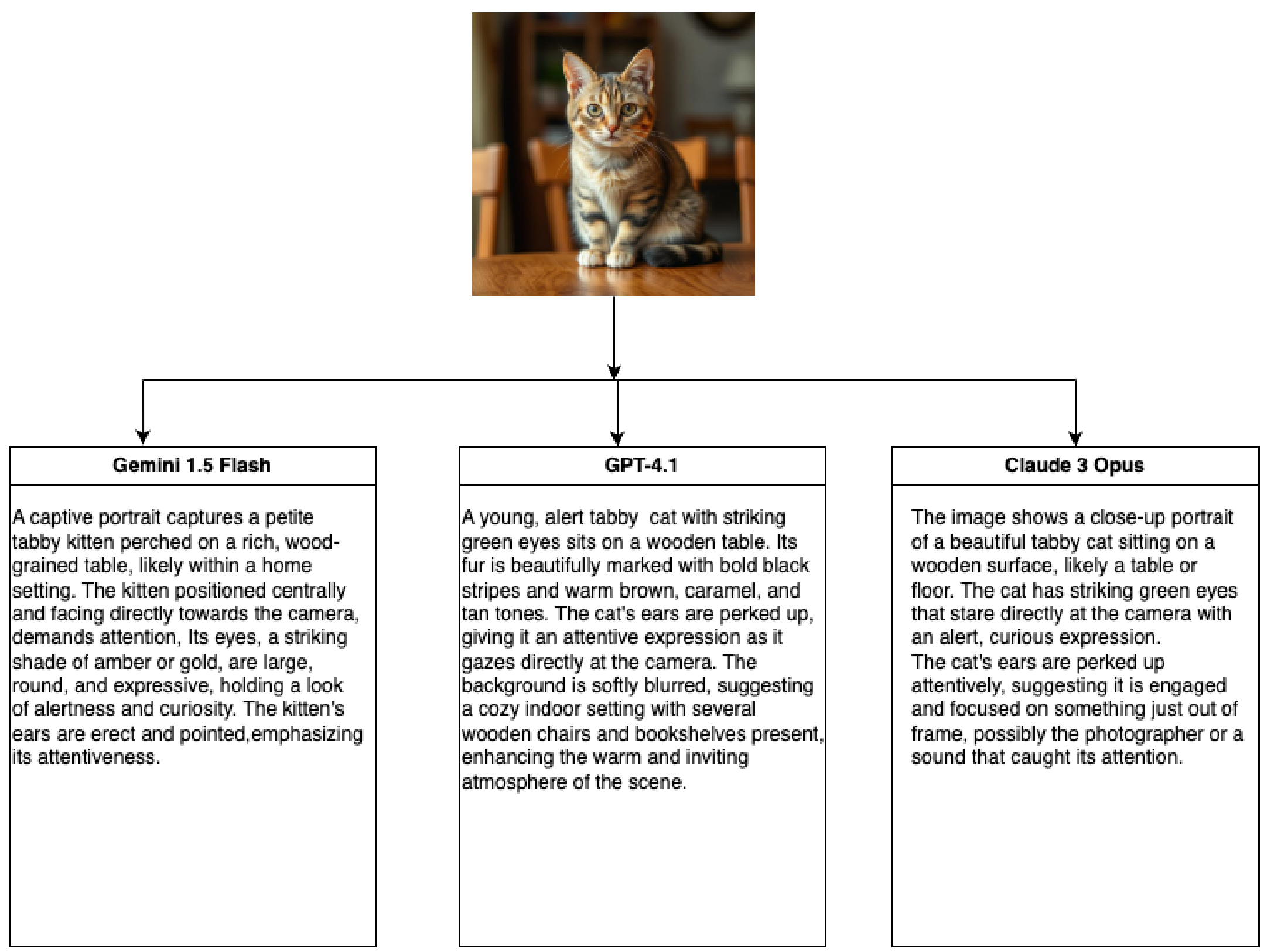

Fig. 2. Comparison of different modern Commonsense integrated Vision Language Models predicting a cat's attention based on the physical expression. While all the models pointed out the cat's attention, the Claude 3 Opus model pointed out the expression in detail, stating the cat's alertness and curiosity.

Figure 2 demonstrates how modern visual language models caption images by integrating observed scenes and background knowledge about these scenes. For example, the models utilized background knowledge about what it means when a cat's ears are perked up. This formulation emphasizes the importance of integrating direct perception with commonsense priors to achieve robust and generalizable inferences. Emerging research directions reveal promising trends. Scene graphs enriched with commonsense triples enhance object relation modeling. Visual Question Answering (VQA) tasks increasingly require models to reason beyond the visible, posing questions that demand external knowledge. Neurosymbolic methods attempt to blend visual representations with logic-based reasoning. Language-augmented vision transformers leverage pretrained language models to provide external priors and bridge the semantic gap between vision and knowledge.

Nevertheless, formidable challenges remain. Efficient retrieval and integration of relevant commonsense facts at inference time, aligning visual and symbolic representations, and designing architectures that reason causally rather than associatively are open research problems. Addressing these challenges is crucial for advancing towards human-like visual intelligence.

This paper provides a systematic survey of the landscape of commonsense reasoning in computer vision. We review knowledge sources, integration methodologies, target applications, evaluation benchmarks, and open challenges. Our objective is to:

- Provide a comprehensive understanding of how commonsense reasoning can be embedded into vision systems.
- Outline future research opportunities in this vital and rapidly developing field.

## 2 What This Survey Adds Beyond Existing Reviews

While several surveys have examined aspects of commonsense reasoning in vision—including visual question answering (VQA) (Antol et al. 2015; Kafle and Kanan 2017), neuro-symbolic integration (Garcez and Lamb 2020; Sarker et al. 2021), and scene graph generation (Johnson et al. 2015; Xu et al. 2017)—this work offers three unique contributions that distinguish it from prior literature.

### 2.1 Unified Pipeline Abstraction Across Paradigms

Existing surveys typically examine individual methodologies (knowledge graphs, vision transformers, retrieval-augmented systems) in isolation. Our primary contribution is the development of a *unified pipeline abstraction* (Table 3) that systematically decomposes commonsense-enhanced vision systems into four canonical stages:

(1) Perception (visual encoding),
(2) Knowledge Retrieval (external commonsense access),
(3) Fusion (integration of visual and symbolic representations),
(4) Reasoning (inference and decision-making).

This abstraction reveals that seemingly disparate approaches—from early neuro-symbolic systems to modern foundation models—share fundamental architectural patterns. By mapping diverse methods onto a common framework, we enable:

- **Cross-paradigm comparison**: identifying which pipeline stage a method optimizes and where bottlenecks arise;
- **Systematic gap analysis**: exposing underexplored combinations (e.g., efficient retrieval mechanisms for real-time segmentation);
- **Design guidance**: enabling practitioners to compose hybrid pipelines by integrating components across paradigms.

To the best of our knowledge, no prior survey provides this level of methodological unification across the full spectrum of commonsense reasoning approaches in vision.

### 2.2 Paradigm Shift Analysis: Static → Dynamic → Implicit Commonsense (2019–2025)

We identify and formalize a fundamental paradigm shift in how commonsense knowledge is integrated into vision systems:

**Pre-2019 (Static Knowledge Era).** Commonsense knowledge was primarily encoded in fixed knowledge bases and scene graphs, such as ConceptNet and ATOMIC, requiring manual curation and explicit symbolic reasoning. Representative systems include NSCL (Mao et al. 2019a) and early Graph R-CNN models (Yang et al. 2018a).

**2019–2021 (Dynamic Retrieval Era).** The emergence of large-scale vision–language models such as CLIP (Radford et al. 2021a) enabled dynamic retrieval of external knowledge. Methods such as KRISP (Marino et al. 2021a) and REVIVE (Y. Lin et al. 2022a) introduced retrieval-augmented pipelines, though integration remained modular and task-specific.

**2022–2025 (Implicit Knowledge Era).** Recent foundation models, including BEiT-3 (W. Wang et al. 2023), Flamingo (Alayrac et al. 2022), and GPT-4V (OpenAI 2023), internalize large amounts of commonsense knowledge during large-scale pretraining. While explicit retrieval is often eliminated, this shift introduces new challenges related to interpretability, factuality, and controllability.

This temporal analysis demonstrates that progress in the field reflects not merely incremental accuracy improvements, but a fundamental architectural transformation in how knowledge is represented, accessed, and integrated. Understanding this shift is essential for:

- evaluating trade-offs between transparency and efficiency,
- anticipating hybrid architectures that combine implicit priors with explicit verification,
- contextualizing benchmark gains achieved by post-2022 models.

### 2.3 Retrieval-Augmented Reasoning as a Structural Innovation

While retrieval-augmented generation (RAG) is well-established in natural language processing (Lewis et al. 2020), its application to vision–language commonsense reasoning constitutes a qualitatively distinct architectural paradigm.

Unlike text-only RAG systems, vision-based retrieval faces several unique challenges:

- **Multimodal grounding**: retrieval queries must be grounded in spatial regions, temporal segments, or abstract visual concepts;
- **Knowledge source heterogeneity**: systems must integrate structured knowledge graphs, unstructured web corpora, and parametric model knowledge;
- **Inference-time adaptability**: retrieval enables access to up-to-date commonsense without retraining, addressing a major limitation of purely parametric models.

Our analysis in Section 6.1 provides the first systematic treatment of retrieval-augmented vision reasoning as a distinct architectural class, examining query formulation strategies, retrieval source selection, fusion mechanisms, and multimodal-specific failure modes such as hallucination amplification and context mismatch.

## 3 Review Methodology

### 3.1 Paper Selection and Categorization

In this survey, the articles selected for review were chosen based on their contributions to integrating commonsense reasoning into computer vision (CV) tasks. The scope was focused on works published between 2018 and 2025, as these represent the latest and most impactful research in the field. Specific inclusion criteria guided the selection to ensure relevance and impact.

First, only papers that explicitly address the incorporation of commonsense knowledge into computer vision models were included. These models typically integrate knowledge from external sources, such as structured knowledge graphs or neuro-symbolic systems that combine neural networks with logical reasoning. This approach ensures that the selected studies are directly related to the intersection of commonsense reasoning and vision tasks.

Furthermore, the papers were evaluated based on their impact in advancing commonsense integration within vision systems. Only those who demonstrated a significant contribution to the state of the art - either through novel methodologies, datasets, or conceptual frameworks - were included in the survey.

Finally, to provide a comprehensive overview of the field, the selected articles were categorized into three broad groups: 1. **Knowledge Graphs and Scene Graphs**: These articles focus on the use of external knowledge sources, such as ConceptNet or ATOMIC, to enhance vision models. 2. **Neuro-Symbolic Models**: These articles explore approaches that combine deep learning with symbolic reasoning to allow the model to reason more effectively about context and relationships. 3. **Commonsense-Enhanced Vision Transformers**: This category includes papers that integrate commonsense reasoning into transformer-based architectures for computer vision tasks.

### 3.2 Evaluation Criteria

Each paper was evaluated based on several critical criteria. The primary evaluation criterion was accuracy, which refers to the degree to which the models perform on standard computer vision tasks, such as object detection, semantic segmentation, and classification, when integrating common sense knowledge. Performance metrics such as mean average precision (mAP) and Intersection over Union (IoU) were used to assess the models' capabilities.

Additionally, the effectiveness of commonsense integration was assessed. This refers to how well the commonsense knowledge is incorporated into the model's decision-making process. Specifically, we examined whether the commonsense knowledge is used dynamically during inference or whether it is pre-encoded into the model's structure.

Another vital factor was scalability. We considered how well each model could handle large-scale, real-world datasets and the complexity of commonsense reasoning in those datasets. Lastly, generalization was a key criterion — we evaluated whether the models could generalize to new, unseen tasks or domains without significant performance degradation.

### 3.3 Methodological Framework

A structured framework was developed to categorize and analyze the selected papers systematically. This framework involves analyzing the approach, effectiveness, and challenges in each category (Knowledge Graphs, Neuro-symbolic Models, and Vision Transformers). Each paper within these categories was reviewed for its methodology, the techniques employed to integrate commonsense reasoning, and its performance based on the evaluation criteria outlined earlier.

The methodological framework is visually represented in Table 1, which illustrates the categorization of the reviewed papers, along with a flow of the evaluation process. This framework was used to ensure a consistent and fair comparison across the different approaches, providing insight into their strengths and weaknesses.

In this section, we have outlined the methodology used to select, categorize, and evaluate the papers included in this survey. The selected papers encompass the most recent advances in integrating commonsense reasoning into computer vision models, classified by the methods employed (knowledge graphs, neuro-symbolic models, and transformers). We have also detailed the criteria used to assess these papers, including accuracy, commonsense integration, scalability, and generalization.

Table 1. PRISMA Flow Diagram of Study Selection Process

| **IDENTIFICATION** |
| --- |
| Records retrieved from: |
| - IEEE Xplore (n=200) |
| - arXiv (n=450) |
| - CVPR/ICCV/ECCV Proceedings (n=250) |
| - Additional Sources (n=150) |
| **Total Records:** 1,050 |
| ↓ |
| **SCREENING** |
| Duplicates Removed: 450 |
| Records Screened (Title/Abstract): 560 |
| Excluded: 400 |
| - Irrelevant (no commonsense focus) (n=287) |
| - Non-Vision Tasks (n=113) |
| ↓ |
| **ELIGIBILITY** |
| Full-Text Articles Assessed: 160 |
| Excluded: 117 (Reasons): |
| - Pre-2018 Studies (n=64) |
| - Insufficient Evaluation (n=20) |
| - Less Significant Contributions (n=28) |
| - Theoretical Papers (n=5) |
| ↓ |
| **INCLUDED** |
| Final Studies: 44 |
| - Foundation Models (n=10) |
| - Vision-Language Integration (n=25) |
| - Surveys/Meta-Analyses (n=9) |

## 4 Commonsense in Computer Vision: Key Areas

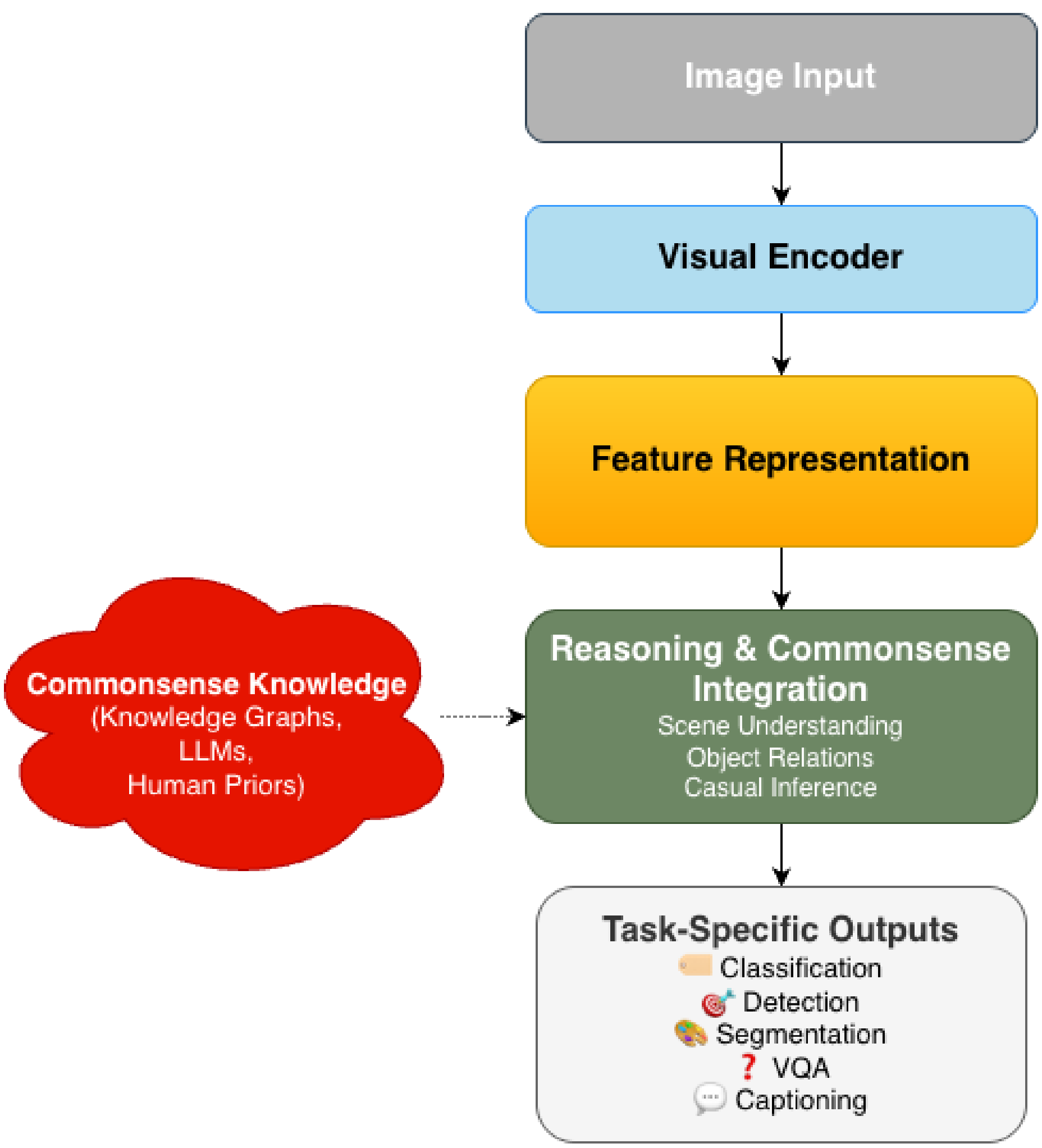

Fig. 3. Conceptual pipeline of commonsense reasoning in computer vision. Commonsense knowledge from external sources (knowledge graphs, large language models, and human priors) enhances the reasoning stage, enriching visual features with contextual, causal, and relational understanding for improved task performance across classification, detection, segmentation, VQA, and image captioning.

While deep learning has led to remarkable advances in computer vision tasks, models often still lack commonsense understanding—reasoning that humans naturally apply when interpreting complex scenes. In this section, we critically examine how commonsense reasoning enhances key vision tasks, including classification, detection, segmentation, visual question answering (VQA), and image captioning Figure 3. Each subsection analyzes the role of commonsense knowledge, presents examples, and highlights challenges that persist.

## 4.1 Image Classification: Contextual Classification

Traditional image classification models typically predict object categories based on local visual features extracted from images, such as color, texture, and shape, using predefined labels by convolutional or transformer backbones (**Chollet2019**). However, in many real-world settings, objects appear in ambiguous, occluded, or novel configurations where isolated visual patterns are insufficient. Contextual image classification aims not only to classify objects in an image ("dog", "car") but also to interpret them in context, considering the scene, relationships, functionality, and affordances. Commonsense reasoning enables classification models to leverage scene context, object co-occurrence priors, and environmental affordances, making more plausible predictions. These models often overlook the importance of spatial context, which can significantly enhance classification accuracy and understanding of scene semantics.

For instance, consider the image in Figure 4 where the visible part of a zebra is occluded, showing only faint stripe patterns. Without understanding that zebras typically appear in grassy savannahs or near herds of similar animals, a model may misclassify the object. Humans, by contrast, use contextual cues such as background vegetation, nearby animals, and typical scene layouts to infer the correct label. In autonomous vehicle perception, context-aware classification is critical for safety. A vision system must distinguish between a plastic bag blowing across the road (which can be safely ignored) and a similar-sized object that poses a collision risk. This requires reasoning about object materials, typical behaviors, and environmental context—knowledge that cannot be reliably learned from pixel patterns alone. Similarly, in industrial quality control, classifying a component as "defective" requires understanding not just visual appearance but typical manufacturing tolerances, functional requirements, and downstream assembly constraint

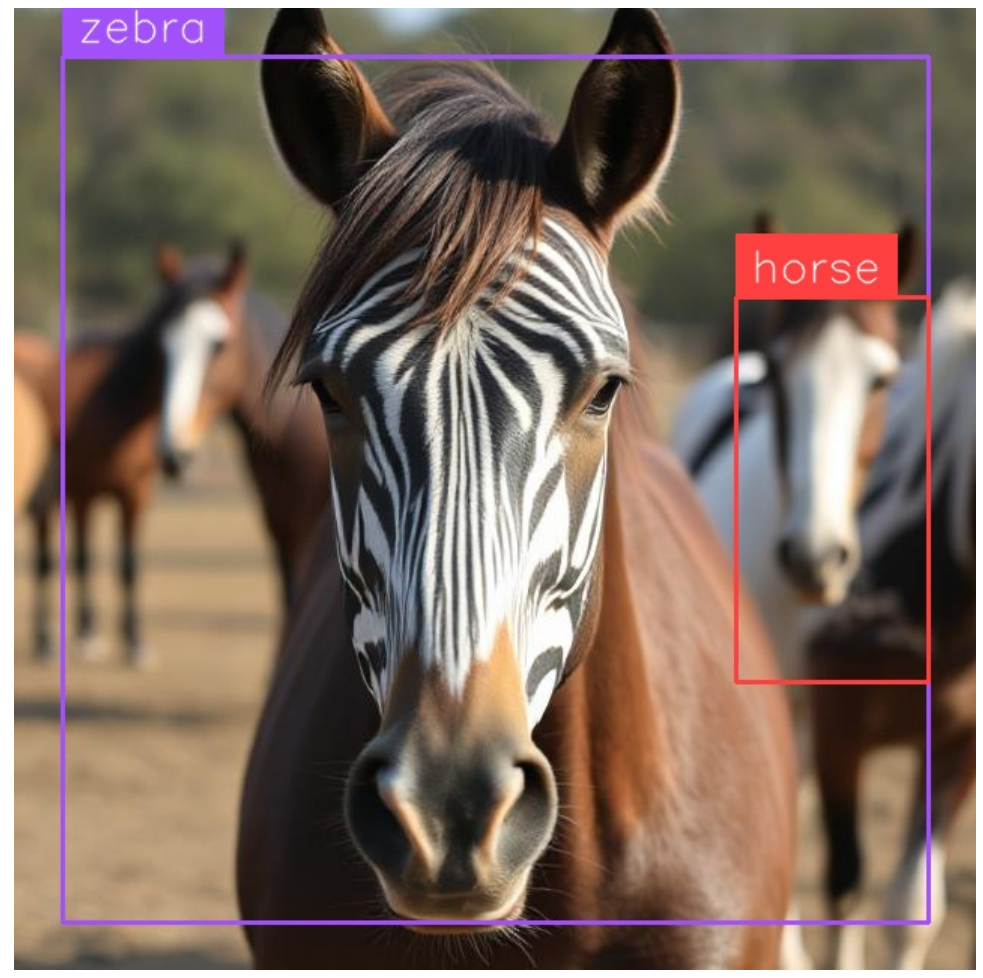

Fig. 4. A horse with some zebra stripes on its face near other horses in a ranch. A model may misclassify this horse as a zebra by just looking at the zebra patterns on its face without using contextual cues such as background vegetation or nearby animals.

Recent datasets such as ImageNet-A demonstrate that even state-of-the-art classifiers fail significantly under out-of-distribution and ambiguous conditions (Hendrycks, K. Zhao, et al. 2019), highlighting the need for context-aware commonsense reasoning in classification models.

### 4.2 Object Detection: Spatial Relations and Affordances

Spatial relations and affordances play a crucial role in visual object detection, enhancing the ability of systems to interpret and interact with their environment in a manner akin to human cognition. Spatial relations refer to the geometric and positional relationships between objects (J. U. Kim et al. 2020) while affordances describe the potential actions that objects offer to an agent. Together, these concepts contribute to the development of more robust and interpretable object detection systems. Traditional object detection methods typically work by creating separate branches for predicting the class and the bounding box of each object. This approach can overlook the spatial and logical relationships between objects, which are crucial for accurate detection (Yang et al. 2024).

For example, detecting a "cup" floating mid-air in a scene rather than placed on a table indicates a failure to apply physical commonsense about support relations. Similarly, detecting a "chair" upside down on a ceiling would normally trigger a contradiction in human reasoning.

Visual commonsense reasoning enhances object detection by leveraging spatial relationships and object affordances, enabling models to understand contextual interactions among objects. This integration enhances localization accuracy and coherence, leading to more rational and contextually relevant detection outcomes. Models enhanced with relational reasoning modules, such as Relation Networks (Hu et al. 2017), can achieve state-of-the-art performance on challenging tasks like visual question answering, where understanding the relationships between objects in an image is essential (Santoro et al. 2017). Integrating knowledge graphs of physical affordances (e.g., cups usually sit on flat surfaces) into detection pipelines remains an open research direction.

Datasets like Visual Genome (Krishna et al. 2017) provide object relation annotations that can serve as supervision for commonsense-aware detectors, yet few models fully exploit them at inference time.

### 4.3 Image Segmentation: Scene Understanding

By leveraging advanced image segmentation techniques, AI systems can achieve a more comprehensive understanding of scenes, enabling them to discern intricate details and contextual elements that traditional methods may overlook. Traditional segmentation models often struggle with complex scenes because they rely on pixel-level information without considering the broader context.

For example, in urban street scenes, segmenting a "road" versus a "sidewalk" purely based on color and texture can be unreliable under varying lighting or weather conditions. Commonsense knowledge about city layouts—such as sidewalks typically bordering roads, crosswalks connecting sidewalks, and roads hosting vehicles—helps disambiguate confusing visual cues.

Integrating commonsense reasoning into segmentation processes can significantly enhance the ability of AI systems to interpret and segment images more effectively. This improved scene understanding is crucial for applications requiring precise recognition of object boundaries and interactions, ultimately enhancing the overall performance of computer vision systems. Furthermore, integrating commonsense reasoning into image segmentation can facilitate more accurate object delineation, thereby enhancing the model's ability to interpret complex scenes effectively.

Approaches like Panoptic Segmentation (Kirillov et al. 2018) aim to significantly enhance scene understanding by integrating both semantic and instance segmentation, but still lack explicit integration of physical and spatial commonsense. Wang, 2021 (C. Wang 2021) recommended that future segmentation models should incorporate priors about plausible object co-occurrence and spatial arrangements to enforce scene consistency. This improvement could involve exploring new techniques for merging these tasks in a way that maximizes the strengths of each other, resulting in a more comprehensive scene understanding.

Scene parsing datasets, such as Cityscapes (Cordts et al. 2016) and ADE20K (Zhou et al. 2017), offer valuable resources for training and evaluating commonsense-guided segmentation methods, ultimately leading to improved performance in complex scene understanding and enhancing the AI's ability to interpret diverse visual contexts.

### 4.4 Visual Question Answering (VQA): Commonsense QA

Visual Question Answering (VQA) is one of the most promising applications of commonsense reasoning in computer vision, as it necessitates not only object recognition but also an understanding of relationships and context to answer questions accurately.

Questions such as "Why is the man holding an umbrella indoors?" cannot be answered merely by detecting objects; they demand reasoning about uncommon scenarios (e.g., indoor rain simulation or photography props). Similarly, "What will happen if the boy lets go of the balloon?" requires predicting unobserved future outcomes based on physical commonsense.

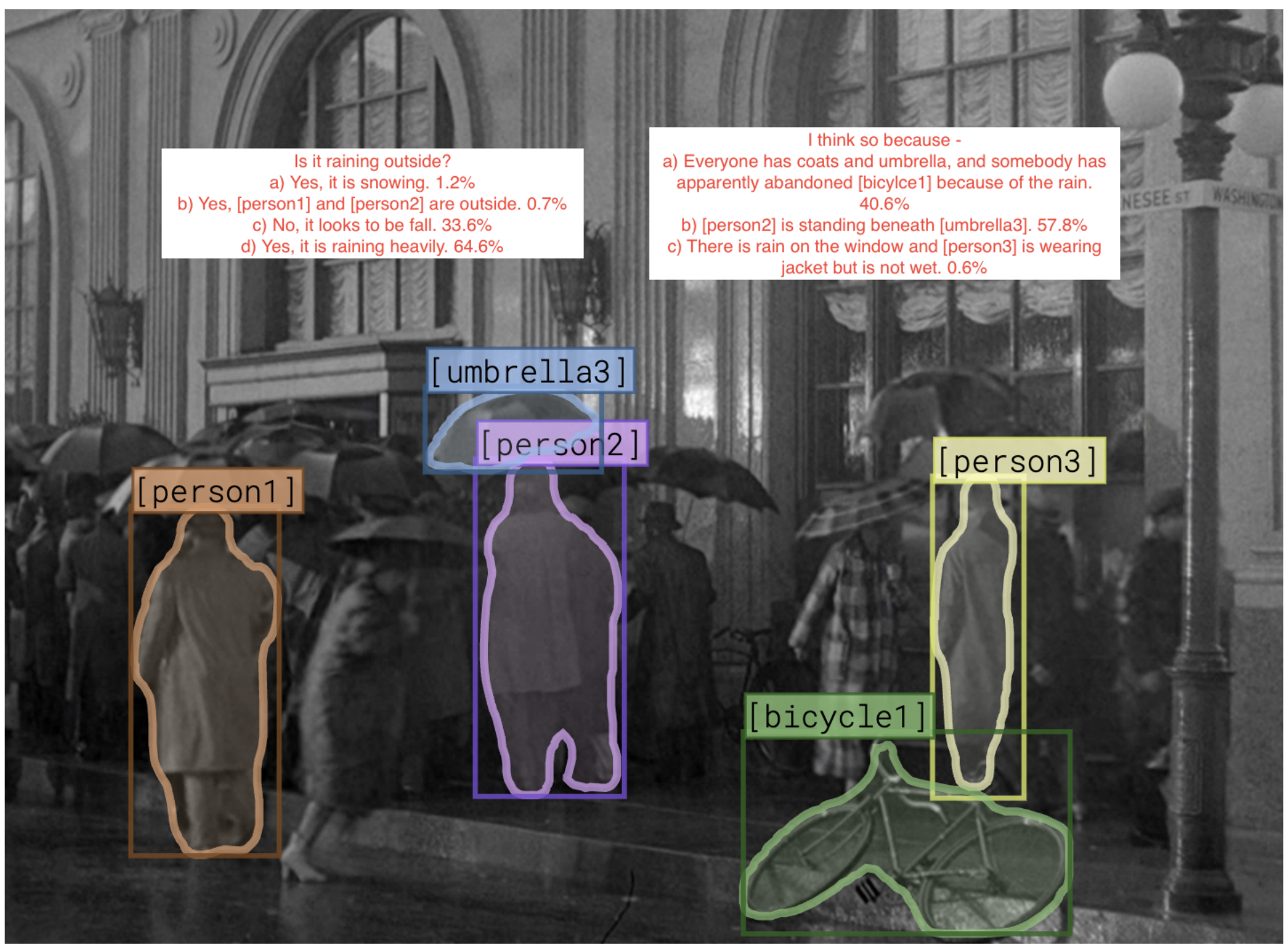

Fig. 5. Application of Visual Question Answering (VQA) in the Visual Commonsense Reasoning (VCR) dataset, where question answering and reasoning are demonstrated. The example shows that the reason people are holding umbrellas is inferred to be rain after answering multiple-choice questions.

Benchmarks like GQA (Hudson and Manning 2019) and VCR (Zellers et al. 2019) introduced structured reasoning demands into VQA evaluation, paving the way for more sophisticated models that can effectively address these challenges. However, even leading models often guess the answer based on superficial correlations rather than true understanding, highlighting the need for improved reasoning mechanisms in VQA systems. Enhancing these models with commonsense knowledge could significantly reduce reliance on such correlations, fostering more reliable interpretations of visual data.

Recent advancements, such as retrieval-augmented VQA models like KRISP (Knowledge Retriever for Inferencing and Semantic Parsing) (Marino et al. 2021b), demonstrate that external knowledge retrieval can improve performance. However, dynamic reasoning over retrieved knowledge remains challenging and requires further exploration to ensure that the models can effectively integrate and utilize this information in real-time scenarios.

### 4.5 Image Captioning: Context-Aware Captions

Commonsense knowledge is crucial for generating meaningful, context-rich image captions that extend beyond merely listing detected objects. Traditional captioning models tend to produce generic templates (e.g., "A person riding a bike") without capturing finer commonsense details like intention, emotion, or scene dynamics.

Context-aware captioning requires models to consider both the visual features and the surrounding textual context, ultimately leading to more accurate and meaningful descriptions of images. For example, a context-rich caption might describe "A child joyfully chasing bubbles in a park," inferring emotion and interaction from the scene.

Large-scale vision-language models like Flamingo (Alayrac et al. 2022) and OmniVL (X. Wang et al. 2023) begin to generate more nuanced captions by pretraining across multimodal corpora. Nevertheless, achieving human-like commonsense depth in generated narratives remains an open challenge, particularly under domain shift or unusual situations.

Datasets like COCO-Captions (T.-Y. Lin et al. 2014) and NoCaps (Agrawal et al. 2018) provide valuable evaluation settings for measuring the commonsense richness of generated descriptions.

## 5 Survey of Commonsense Methods in Vision

### 5.1 Knowledge Graphs and Scene Graphs

The integration of commonsense reasoning in computer vision has seen substantial growth with the use of knowledge graphs (KGs) and scene graphs (SGs), allowing machines to go beyond perception and perform semantic understanding and reasoning over visual data. This subsection examines how KGs and SGs are utilized to enhance the comprehension of visual data and how they contribute to bridging the gap between perception and commonsense reasoning.

*5.1.1 Knowledge Graphs: Contextualizing Visual Data.* A knowledge graph provides a rich, hierarchical structure of concepts, their relationships, and their attributes, offering a more comprehensive understanding of the world. It is a structured database that stores facts as triples: [Subject, Predicate, Object] (X. Jiang et al. 2023). For instance, an external knowledge graph such as ConceptNet encodes relationships like "[cats are animals]" or "[dogs are friendly]", which are essential for reasoning about objects in a scene.

To understand the workflow of a knowledge graph, in Fig 6, a cat is sitting next to a half-full bowl. A traditional computer vision model, such as YOLO, will detect the objects in the image, where the detected objects are [Cat, Bowl, White Liquid]. After querying the KG with the detected objects:
Query: "Cat" → Returns: <Cat, eats, Food>, <Cat, drinks, Milk>
Query: "Bowl" → Returns: <Bowl, usedFor, Holding Liquids>

An inference after combining Knowledge Graph (KG) and Computer Vision (CV) could be

- Observation: Cat + Bowl + White liquid.
- KG Fact: Cats drink milk.

- Inference: The liquid is likely milk → "Cat is drinking milk."

Table 2 gives an example, how KG can enhance CV tasks.

Table 2. Comparison of Image Captioning and Visual Question Answering with and without Knowledge Graph (KG) augmentation

| **Task** | **Without KG** | **With KG** |
|---|---|---|
| **Image Captioning** | "A cat and a bowl." | "A cat drinks milk from a bowl." |
| **Visual Question Answering** | "Is the cat thirsty?" → Guess | "Yes" (KG: Cats drink milk → thirst implied) |

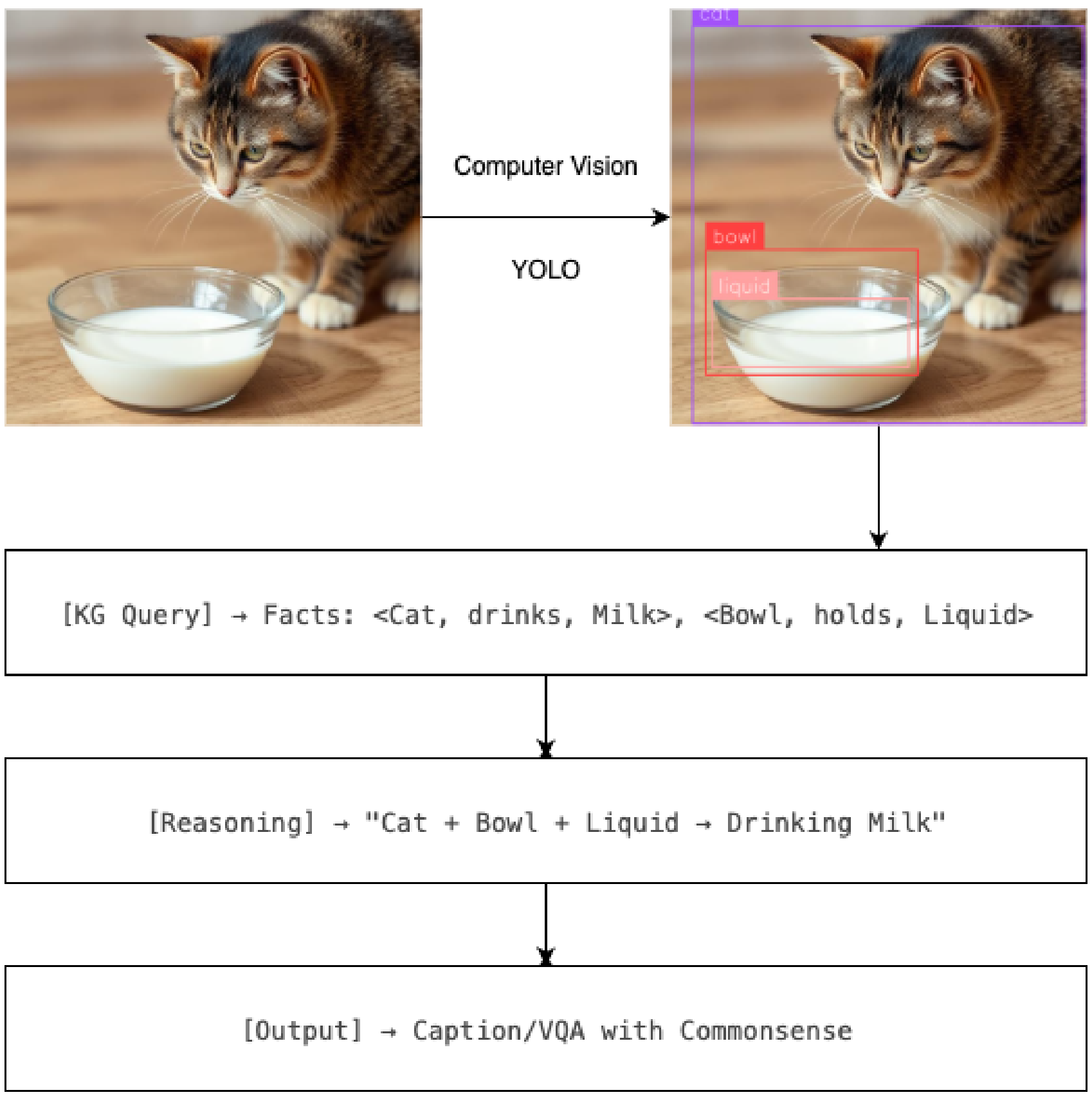

Fig. 6. Flowchart illustrating how Knowledge Graphs (KGs) integrate with Computer Vision (CV) systems for enhanced reasoning and contextual understanding.

Recent advancements in integrating Knowledge Graphs (KGs) with deep learning models have demonstrated that external knowledge can significantly improve a model's ability to reason about objects and their interactions by providing contextual and commonsense information (Mao et al. 2019a; Marino et al. 2021b; Speer et al. 2017a).

These approaches support reasoning over entities, actions, and relationships while grounding visual predictions in world knowledge.

For example, KRISP integrates symbolic knowledge graphs with neural visual reasoning to improve performance in knowledge-based visual question answering tasks (Marino et al. 2021b). Similarly, the Neuro-Symbolic Concept Learner (NSCL) combines neural perception with symbolic reasoning to enable interpretable visual reasoning (Mao et al. 2019a). Such systems allow models to infer commonsense relationships, such as recognizing that a cat on a table is more likely sitting rather than floating, based on learned semantic and relational knowledge.

Furthermore, neural-symbolic integration has made significant progress in combining knowledge graphs with neural architectures to enable logical and explainable reasoning. However, a major challenge remains the dynamic integration of relevant commonsense knowledge during inference, particularly in real-time scenarios where contextual information is incomplete or ambiguous.

Furthermore, neural-symbolic integration has made significant strides in combining knowledge graphs with neural networks to enable logical reasoning. The key challenge, however, lies in the dynamic integration of such knowledge in real-time systems, where the system needs to retrieve relevant commonsense knowledge during inference – common in real-world scenarios in which context is not fully provided in the input.

*5.1.2 Scene Graphs: Modeling Object Interactions and Relations.* A scene graph extends the concept of a knowledge graph by incorporating not only objects and their relationships but also their spatial arrangements and attributes. Scene graphs are instrumental in image captioning, object detection, and visual reasoning tasks, where understanding the interactions between objects is key. In contrast to traditional object detection models, which identify individual objects without considering their relationships, scene graphs facilitate a deeper understanding of semantics.

Scene graphs represent objects in an image as nodes and relationships between them as edges, forming a graph that encodes both spatial and functional relationships. For example, in a typical picture with a dog and a ball, a scene graph might represent the relationship as "dog → playing with → ball," helping the model to understand not just the presence of objects but their contextual interactions.

To understand the workflow of the Scene Graph augmented Computer Vision, let's take an example of the image in Figure 7 , where a girl is kicking a soccer ball in a park. A traditional computer vision model will detect objects where the detected objects are [girl, soccer ball, grass]. After object detection, the scene graph model identifies relationships and attributes between objects, such as (girl → kicking → soccer ball), (soccer ball → on → grass). Then the model will provide a caption based on the identified relations between objects ("A girl kicks a soccer ball on grassy ground.").

Now, we can see how SG has enhanced CV tasks from the following Table 3:

Table 3. Image Captioning and Visual Question Answering with and without KG augmentation

| **Task** | **Without SG** | **With SG** |
|---|---|---|
| **Image Captioning** | "A girl and a soccer ball." | "A girl kicks a soccer ball on grassy ground." |
| **Visual QA** | "What is the girl doing?" → "Standing." (guesses from object detection) | "Kicking a soccer ball." (from girl → kicking → soccer ball) |
| **Image Search** | Finds all images of people playing outside | kicking (action) + grass (outdoor) → Higher accuracy. |

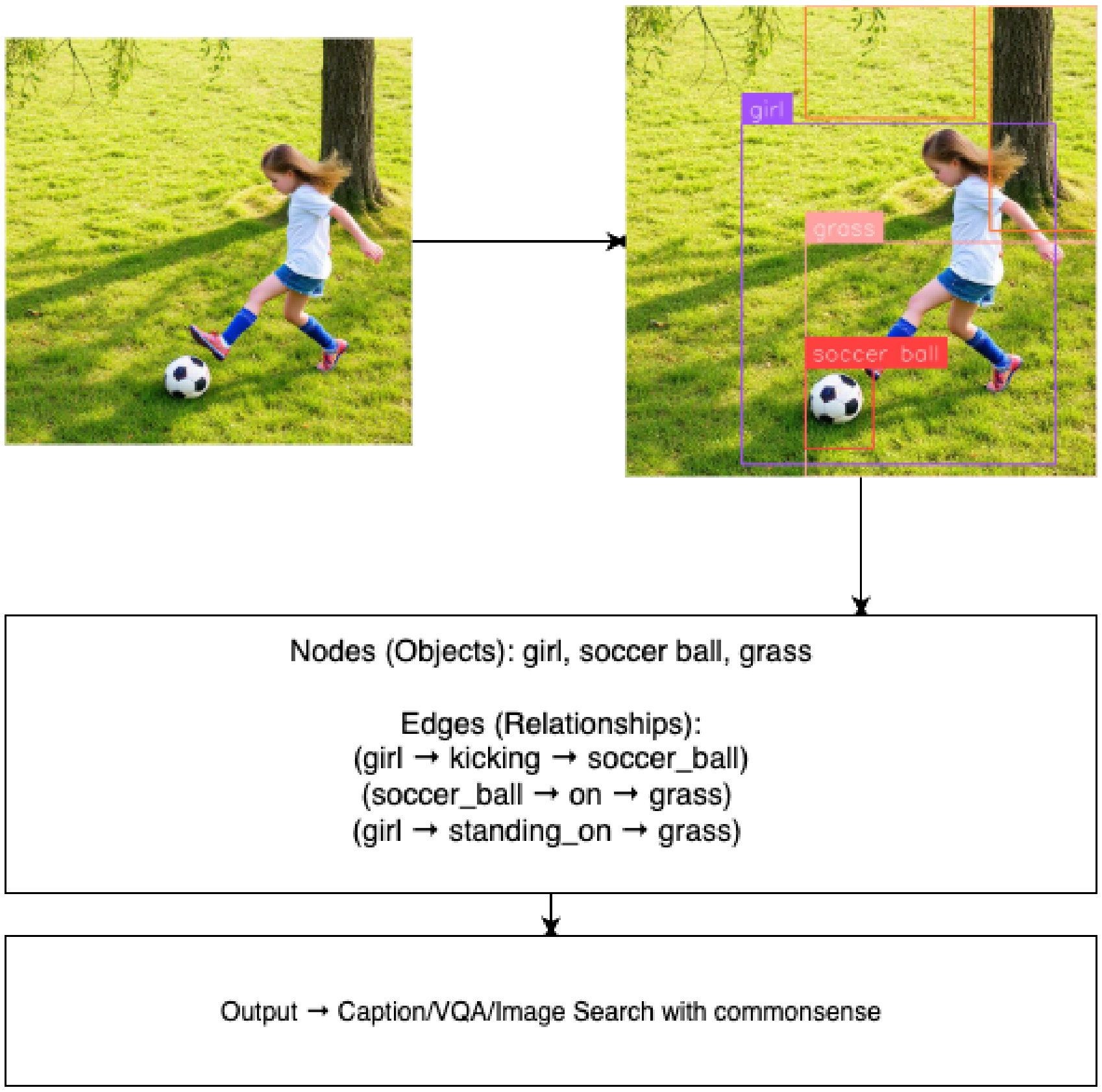

Fig. 7. Flowchart illustrating how Scene Graphs (SGs) integrate with Computer Vision (CV) systems for enhanced relational reasoning and contextual understanding.

In robotics manipulation tasks, understanding object affordances—the action possibilities offered by objects—is essential. A robotic system must recognize not only that an object is a "cup" but also that it can be grasped by the handle, filled with liquid, and placed upright on a flat surface. This requires commonsense knowledge about object functions, typical usage patterns, and physical constraints. In industrial inspection scenarios, detecting defects requires understanding what constitutes "normal" spatial arrangements. For example, identifying misaligned components on an assembly line requires knowledge of correct positioning, typical tolerances, and functional relationships between parts.

By embedding scene graphs into the network architecture, such as through graph convolutional networks (GCNs), researchers have significantly improved image understanding. Models like Graph R-CNN (Yang et al. 2018b) and Relation Networks have pushed the boundary by allowing networks to not only detect objects but also understand their relations within a scene, based on a deeper understanding of how objects relate to one another in the real world. This integration has enabled better handling of complex scenes, where understanding object relationships is crucial for comprehending the context.

In recent years, transformer-based architectures have been increasingly applied to scene graphs, offering new methods for integrating relational data. For example, the ViLBERT model (Lu et al. 2019), which combines vision and language, uses transformers to generate scene graphs and integrate them with natural language representations. This integration enhances the model's ability to generate context-aware captions and answer complex visual questions, further underscoring the potential of scene graphs to enhance commonsense reasoning in visual tasks.

One significant challenge with scene graphs is scalability. As the number of objects in a scene increases, the complexity of generating and processing scene graphs also increases, resulting in inefficiency. Recent work such as Vision Transformers and DETR (Detection Transformer) (Li et al. 2021) has aimed to combine transformer architectures with scene graph generation to address this issue. These models utilize self-attention mechanisms to focus on relevant object relationships within large-scale scenes, thereby improving performance while maintaining computational efficiency.

However, integrating scene graphs into real-time systems remains a monumental task. The need for dynamic graph generation during inference, particularly in dynamic and cluttered environments, is a significant bottleneck. Furthermore, while scene graphs excel at modeling object relationships, they still struggle with higher-order reasoning, such as causal inference or long-term dependencies in dynamic scenes, which limits their effectiveness in more complex tasks like autonomous driving or interactive robotics.

Additionally, data sparsity in scene graphs remains a concern. While large datasets, such as Visual Genome (Krishna et al. 2017) and COCO (T.-Y. Lin et al. 2014), provide ample data for training scene graph models, there are still many scenarios where scene graphs are incomplete or lack key relationships. Lack of exhaustive relational data can lead to biases or incorrect inferences by the model. A study by Goel et al. (2021) (Goel et al. 2021) highlighted that current scene graph methods, which are trained on the entire set of relations, struggle to develop complex reasoning about visual and textual correlations, primarily due to biases present in the training data.

### 5.2 Neuro-Symbolic Models

Neuro-symbolic models represent a promising direction in commonsense reasoning, as they combine the strengths of neural networks and symbolic reasoning to improve AI's interpretative capabilities in complex visual contexts. These models are designed to leverage the strengths of both approaches: the perceptual power of deep learning for handling raw data and the logical reasoning capabilities of symbolic systems for structured, abstract thinking. In the realm of commonsense reasoning, this hybrid approach has shown promising results by enabling systems to reason for high-level relationships, causality, and implicit knowledge. For example, in Fig.8, a child is reaching toward a knife on a kitchen counter. After object detection ("child", "knife", "counter") and relationship analysis [("child", "reaching", "knife"), ("knife", "on", "counter")] through deep learning based vision model such as Faster R-CNN, the neuro-symbolic model integrates symbolic knowledge such as knowledge graph: Knife[is] → dangerous[unsafe for] → child[requires] → intervention. Then the model performs logical reasoning through several inference steps such as, reaching(child, knife) + dangerous(knife) → unsafe_for(child, knife) → requires_intervention. After that, the model provides actionable output: "A child is dangerously reaching for a knife on the counter.".

Early models focused on integrating traditional symbolic knowledge, like rules and logical expressions, into deep learning pipelines. However, recent advancements have refined this integration, enabling end-to-end learning while preserving the benefits of symbolic reasoning. One of the most notable approaches is the *Neuro-Symbolic Concept Learner (NSCL)* (Mao et al. 2019b), introduced by Mao et al. in (2019). The NSCL combines deep convolutional networks for visual perception with logical reasoning through symbolic representations. This model was able to reason about images by decomposing them into concepts and their relationships, and then using logical rules to infer new facts. This approach brought a significant improvement in tasks that required compositional reasoning, such as visual question answering (VQA). The DeepMind team has also contributed to this area with their work on *DeepLogic* (**deepLogic**), which integrates symbolic reasoning for decision-making in visual tasks. By utilizing a symbolic knowledge base in conjunction with learned neural representations, the model can reason about relationships between objects and infer causal relationships, a critical component of commonsense reasoning.

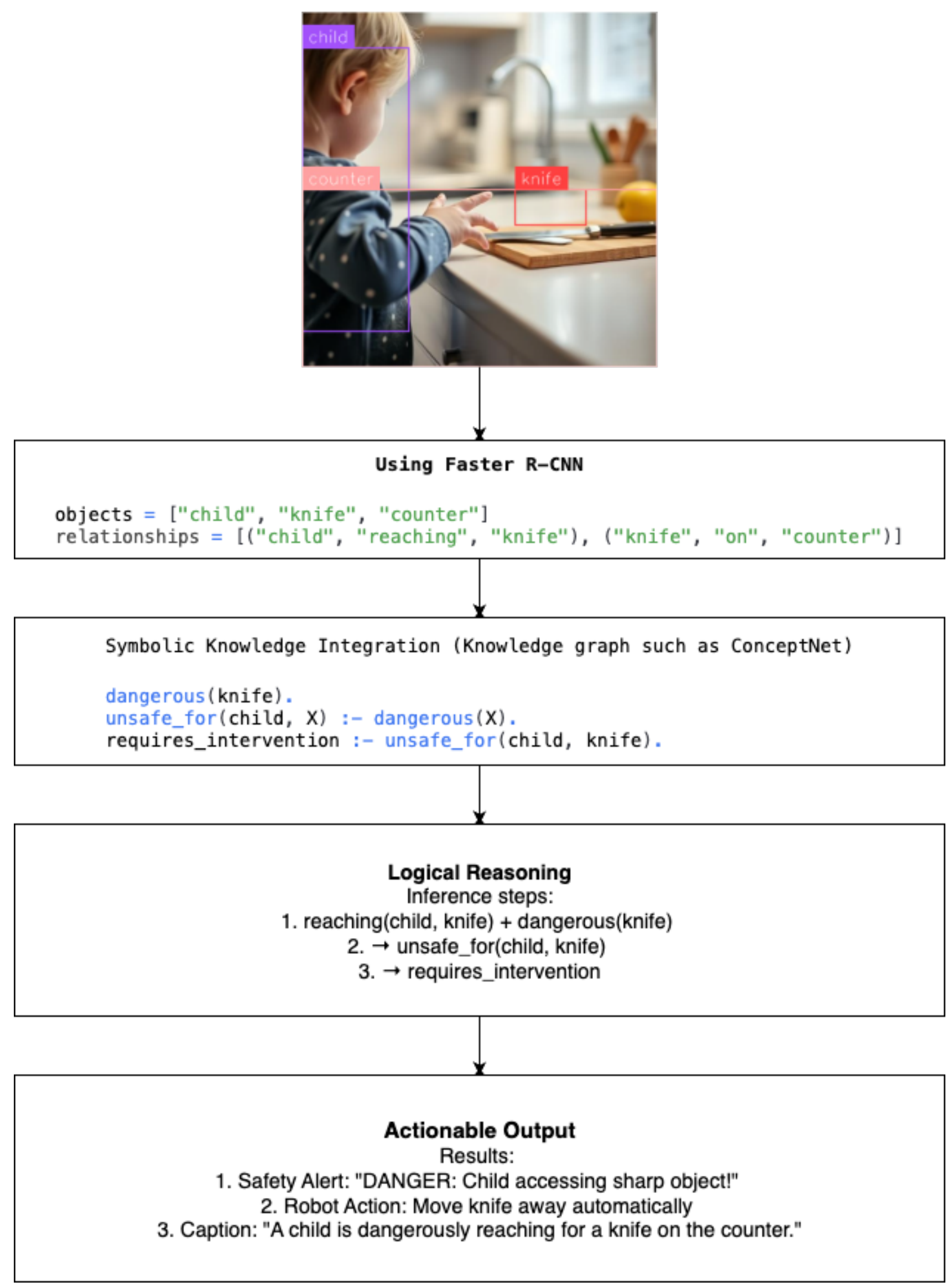

Fig. 8. Flowchart illustrating how neuro-symbolic models combine neural perception with symbolic reasoning to generate actions based on logical rules and structured expressions.

*5.2.1 Neuro-Symbolic Integration in Commonsense Reasoning.* One of the main challenges of commonsense reasoning is how to effectively represent and use abstract, relational knowledge that goes beyond what can be directly observed in the data. While deep learning models excel at learning patterns from data, they often fail at reasoning about abstract concepts, such as intentions or affordances. For example, while a traditional object detection model might identify a knife in an image, it would struggle to understand that the knife is likely to be used for cutting, and not for a different purpose.

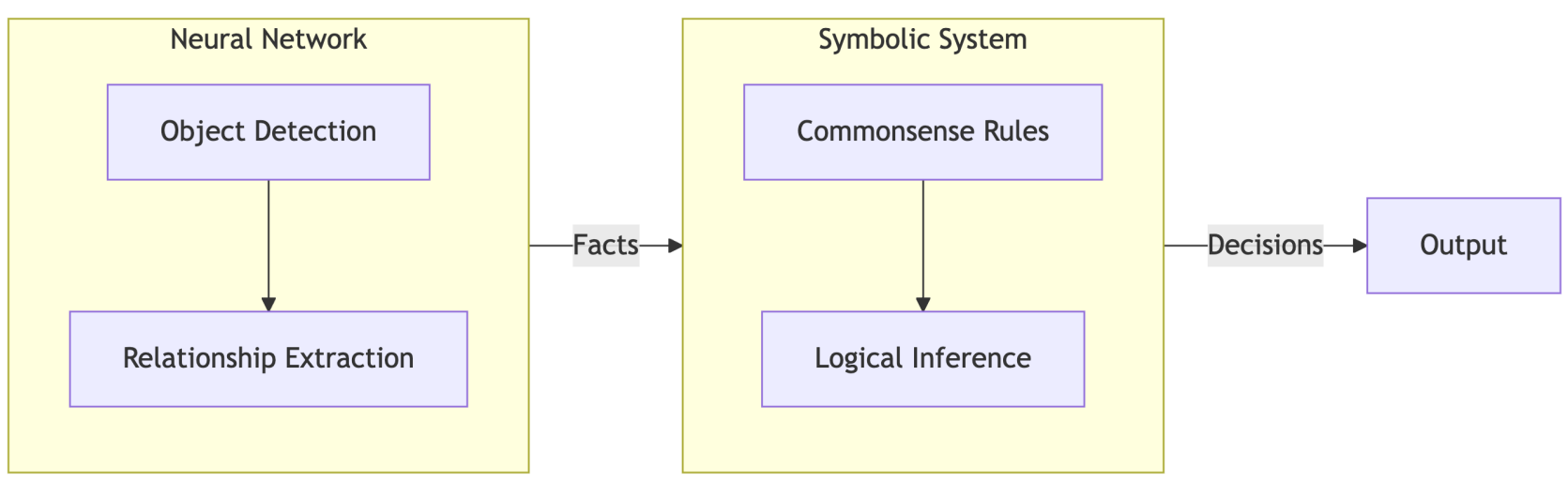

Fig. 9. Graphical representation of a neuro-symbolic pipeline integrating neural networks with symbolic reasoning components.

Recent neuro-symbolic approaches address this challenge by embedding commonsense knowledge directly into the reasoning process. For instance, the work by Zhu et al. (2020) on neuro-symbolic reasoning for robotic manipulation (Zhu et al. 2020) demonstrated how a robotic system can utilize symbolic reasoning to predict possible actions for a set of objects in a scene. By utilizing commonsense knowledge bases such as ATOMIC (**atomic**), these models can infer the intentions of humans and the affordances of objects, enabling the system to perform tasks more effectively.

Moreover, neuro-symbolic models are beginning to show potential in few-shot learning. By combining the generalization power of symbolic reasoning with the data-efficient learning capabilities of neural networks, these models can infer relationships or make predictions with limited examples.

Despite the promise of neuro-symbolic systems, several challenges remain in their development. One of the key limitations is the difficulty in effectively combining the flexibility of neural networks with the rigidity of symbolic systems. While neural networks are highly adaptable and capable of learning from data, symbolic systems often rely on predefined logic and rules, which may not always align with the flexibility required for real-world scenarios. This dissonance between the two paradigms can lead to inconsistent reasoning or limitations in generalization when applied to complex, dynamic environments.

Furthermore, the scalability of neuro-symbolic models remains a significant challenge. While small-scale experiments have shown promise, scaling these models to handle real-world, large-scale data — especially with diverse commonsense knowledge bases — is still an ongoing research problem. The querying and retrieval of relevant symbolic knowledge during inference, particularly in real-time scenarios, remains a bottleneck for the broader adoption of these models in practical applications.

Moreover, integrating commonsense reasoning with symbolic systems is also limited by the incompleteness of knowledge bases. While knowledge graphs such as ATOMIC or ConceptNet provide a large set of relations, they still have gaps, particularly in areas that require dynamic reasoning or complex, multi-step inferences. For instance, many real-world tasks require understanding subtle contextual clues that are not always captured by existing commonsense knowledge bases.

## 5.3 Commonsense-Enhanced Vision Transformers

Vision Transformers (ViTs) have revolutionized the field of computer vision by moving away from traditional convolutional neural networks (CNNs) to transformer-based architectures. ViTs are capable of capturing long-range dependencies in images, making them highly effective for tasks such as image classification, object detection, and segmentation.

*5.3.1 The Role of Commonsense Knowledge in Vision Transformers.* Incorporating commonsense knowledge into ViTs can take several forms, such as leveraging knowledge graphs, knowledge bases, or pretrained language models. The primary objective is to enable the model to not only learn visual representations from the data but also incorporate contextual knowledge that allows for reasoning about the plausibility and interactions of objects within the scene. Commonsense-enhanced ViTs can be used to predict better affordances, object relationships, and intended actions, tasks that require an understanding of how objects behave in the real world.

Recent work in this area includes the use of multi-modal transformers that combine both visual and language representations. LXMERT (Learning Cross-Modality Encoder Representations from Transformers) (Tan and Bansal 2019a) and ViLT (Vision Transformers for Language Understanding) (W. Kim et al. 2021a) are prominent models that integrate language-based commonsense knowledge with visual understanding. These models are trained on multi-modal datasets and can reason about the relationships between visual inputs (images) and textual descriptions, which significantly enhances their commonsense reasoning capabilities.

A key innovation in commonsense-enhanced ViTs is the integration of pretrained language models (PLMs) like BERT (Devlin et al. 2019b) or GPT (Brown et al. 2020), which have been shown to capture rich commonsense knowledge during training. For instance, ViLT uses transformers not only for vision tasks but also for language tasks such as image captioning and Visual Question Answering (VQA). By pretraining the transformer on vast amounts of language data, ViLT learns to associate visual representations with textual representations and commonsense knowledge, enabling the model to infer more plausible relationships and object interactions in images.

Another approach has been augmenting the ViT architecture with commonsense reasoning modules that query structured knowledge bases. For example, a recent work by Zellers et al. (2021) (Zellers et al. 2019) introduced a commonsense-augmented transformer for image captioning tasks, where the model dynamically retrieves relevant knowledge from a commonsense knowledge base such as ATOMIC and uses this information to improve the coherence and context of the captions generated for images. This approach demonstrated that injecting structured knowledge into ViTs during the inference phase can lead to more contextually accurate and commonsense-consistent outputs.

While the integration of commonsense knowledge into ViTs has shown significant promise, several challenges remain. One of the primary obstacles is knowledge retrieval: models must be able to query external knowledge bases efficiently during inference, without introducing significant delays or computational overhead. The dynamic nature of commonsense knowledge — which varies depending on the context, object relationships, and task — presents additional challenges in selecting the most relevant expertise for each task.

Additionally, data sparsity in knowledge bases remains a critical issue. While large-scale knowledge bases cover a wide range of relationships and facts, they still lack comprehensive coverage in many domains, particularly those that require high-order reasoning or domain-specific knowledge. This knowledge gap can lead to models making incorrect or biased inferences, particularly in unfamiliar or edge-case scenarios. Furthermore, integrating commonsense reasoning often requires aligning symbolic and visual representations, a task that has proven difficult due to the differences in how knowledge is represented in each modality.

Another significant challenge is the scalability of commonsense-enhanced ViTs. As the size of the model increases and the complexity of the task grows, the ability to incorporate commonsense reasoning without overfitting or losing generalization capability becomes crucial. Future work will need to explore methods for improving the efficiency of commonsense retrieval and reasoning, especially in real-time applications such as autonomous driving, robotics, and interactive AI systems.

The future of commonsense-enhanced ViTs lies in further integrating multi-modal learning, where vision, language, and commonsense knowledge are combined in a more seamless and coherent way. Models such as CLIP (Contrastive Language-Image Pre-training) (Radford et al. 2021b) have demonstrated the effectiveness of joint training for vision and language. Further research into such frameworks may lead to significant improvements in commonsense reasoning. Additionally, exploring meta-learning techniques for commonsense knowledge could help ViTs generalize better across different domains, thereby reducing their dependency on large, handcrafted knowledge bases.

Another promising direction is the use of neuro-symbolic reasoning in conjunction with ViTs, where a symbolic reasoning layer is added on top of the visual and linguistic layers to enable higher-order inferences, such as causal reasoning or long-term dependencies. This approach can significantly enhance the commonsense understanding of ViTs, allowing them to reason about the intentions, goals, and consequences of actions and events in a scene, which is crucial for more complex visual tasks such as robotic manipulation and autonomous driving.

### 5.4 Commonsense-Enhanced Vision Pipelines

Commonsense reasoning has recently emerged as a complementary layer in computer vision (CV), enabling models to reason beyond perceptual cues by incorporating contextual or causal knowledge. Integrating such reasoning requires a systematic workflow that connects perception, knowledge retrieval, and logical reasoning into a unified architecture. This subsection summarizes how commonsense reasoning can be integrated into representative CV tasks. We decompose each task into its main subtasks and list the key techniques, available tools, and typical evaluation metrics that support practical implementation. This summary provides a concise view of the current state-of-the-art while highlighting reusable design elements that can guide the development of future commonsense-enhanced vision systems.

As summarized in Table 4, a general architectural workflow can be identified across all commonsense-enhanced vision tasks. The process typically begins with the *perception* stage, where visual features are extracted from raw inputs using convolutional or transformer-based backbones. This is followed by *grounding*, which links detected entities or regions to semantic concepts and contextual cues. In the subsequent *knowledge retrieval* phase, external information from structured resources such as ConceptNet, ATOMIC, or Wikidata is incorporated to provide contextual or causal background. The retrieved knowledge is then integrated with visual representations during the *fusion and reasoning* stage, often through neural–symbolic models, prompt conditioning, or adapter-based fusion layers. The *constraint validation* stage verifies that the resulting inferences satisfy logical, spatial, or physical plausibility. Finally, the *prediction and explanation* stage produces interpretable outputs that can be supported by both perceptual evidence and external knowledge. Together, these stages define a modular and interpretable pattern that systematically combines perception, reasoning, and knowledge integration within a unified pipeline.

Table 4. Commonsense-enhanced vision pipelines showing core subtasks, representative techniques, tools, and evaluation metrics.

| Task | Subtasks | Techniques | Tools / Resources | Metrics |
|---|---|---|---|---|
| **Image Classification (context-aware)** | 1) Perception; 2) Scene/context encoding; 3) Knowledge retrieval; 4) Fusion & inference | ViT/Conv backbones with contextual fusion; retrieval-augmented priors | Backbones: ViT, ConvNeXt. Context: CLIP features. Knowledge: ConceptNet, ATOMIC, Wikidata via FAISS/ColBERT. Fusion: FiLM/adapters or prompt-conditioning in VLMs (BLIP-2, LLaVA). | Top-1/Top-5 accuracy; ImageNet-A OOD; ECE |
| **Object Detection (relations & affordances)** | 1) Detection; 2) Relation parsing; 3) Physical/affordance checks; 4) Consistency repair | DETR/YOLO for boxes; scene graphs; physics priors (rule- or learning-based) | Detectors: DETR, RT-DETR, YOLOv8. SG: SceneGraph-Benchmark, MotifNet, PyG/DGL. Affordances: HOI heads (HICO-DET), KG-based support relations. | mAP; Rel-mAP; constraint violations |
| **Semantic/Panoptic Segmentation (scene consistency)** | 1) Masking; 2) Layout priors; 3) Knowledge-guided refinement | Mask2Former, SegFormer, DeepLab; co-occurrence/placement priors; CRF/graph refinement | Datasets: Cityscapes, ADE20K. Refinement: CRF/Graph-CRF; KG-guided label smoothing. | mIoU; PQ; boundary F-score |
| **Visual Question Answering (commonsense QA)** | 1) Grounding; 2) Question parsing; 3) Knowledge retrieval; 4) Reasoning and justification | Region grounding + retrieval-augmented generation (RAG) over knowledge graphs or text; neuro-symbolic reasoning | Grounders: OWL-ViT, CLIP. RAG: KRISP-style retriever, REVIVE lookup; FAISS + Wikidata/ATOMIC. Reasoners: program executors, VLM T-modules. | VQA accuracy; justification fidelity; retrieval precision |
| **Image Captioning (context & intent)** | 1) Object/action cues; 2) Prior injection; 3) Hallucination control | Vision–language decoders with object tags; retrieval-augmented prompting; factuality-aware decoding | Models: BLIP-2, InstructBLIP, LLaVA. Tags: OSCAR++-style. Validation: CLIP factuality filters; constraint decoding. | CIDEr; SPICE; hallucination rate |

Although recent progress demonstrates strong potential, several challenges remain for future research on commonsense-enhanced vision. A key issue is achieving efficient and low-latency retrieval of external knowledge so that reasoning can operate effectively during real-time inference. Another open problem involves the differentiable integration of symbolic or logical constraints into neural architectures without sacrificing scalability or generalization. The dependability of retrieved knowledge also requires further attention, since existing databases may contain noise, incompleteness, or bias that affect model decisions. In addition, most current approaches focus on static images; extending these frameworks to dynamic and temporally consistent video analysis demands new mechanisms for temporal reasoning and multi-frame knowledge consistency. Addressing these challenges will advance the seamless coupling of perception and reasoning in future computer vision systems.

## 6 Comparative Analysis

Commonsense reasoning in computer vision has evolved through various modeling paradigms, including knowledge graph enrichment, neuro-symbolic learning, multimodal transformers, and graph-based relational reasoning. In this section, we perform a detailed comparative analysis of recent methods across several critical dimensions: accuracy, commonsense integration, scalability, and real-world applicability. Table 5, The comparative ratings reported (e.g., *High*, *Medium*, *Low*) are intended as qualitative and relative indicators rather than absolute or directly reproducible measurements. These labels are derived from a synthesis of reported benchmark trends, architectural characteristics, and discussion of deployment constraints in the original publications. Due to differences in datasets, evaluation protocols, and task scope across studies, the table should be interpreted as a high-level comparison of design trade-offs rather than a definitive ranking of methods.

## 6.1 Evaluation Criteria

The models are evaluated based on four key criteria:

- **Accuracy**: Performance on standard vision benchmarks, such as ImageNet and COCO, or task-specific datasets for VQA and captioning.
- **Commonsense Integration**: The depth and method of external commonsense knowledge incorporation, such as knowledge graphs, symbolic reasoning, or pretrained language models.
- **Scalability**: The ability to handle large-scale data, real-time inference, and complex relational reasoning under computational constraints.
- **Real-World Applicability**: Robustness and practicality for dynamic, unseen environments, including autonomous vehicles, robotics, and interactive AI.

## 6.2 Tabular Summary

| Model | Accuracy (%) | Commonsense Integration | Scalability | Real-World Applicability |
|---|---|---|---|---|
| SENet (**senet2020**) | High | High | Medium | High |
| NSCL (Mao et al. 2019b) | Medium-High | Very High | Low | Medium |
| LXMERT (Tan and Bansal 2019b) | High | Medium | High | Very High |
| ViLT (W. Kim et al. 2021b) | Very High | High | Medium-High | Very High |
| Graph R-CNN (Yang et al. 2018b) | Medium | Medium-High | Low | Medium |

Table 5. Comparative analysis of commonsense-enhanced models. Ratings are relative and derived from recent experimental benchmarks and practical deployments.

## 6.3 Critical Discussion

*6.3.1 SENet (Semantic Enrichment Network).* SENet enhances visual representations by incorporating external semantic information from knowledge graphs such as ConceptNet (Speer et al. 2017a). It achieves high accuracy in object detection and segmentation tasks by disambiguating visual features with contextual knowledge. However, the model's reliance on static external graphs can introduce latency during inference, affecting scalability in real-time systems. Despite these limitations, SENet demonstrates strong applicability in medical imaging, autonomous vehicles, and assistive technologies where contextual awareness is critical.

*6.3.2 NSCL (Neuro-Symbolic Concept Learner).* The Neuro-Symbolic Concept Learner (NSCL) integrates perceptual modules (deep convolutional networks) with symbolic reasoning engines, enabling compositional and relational reasoning over visual scenes (Mao et al. 2019b). NSCL excels in tasks requiring structured logical inference, achieving superior commonsense integration. Nevertheless, the model struggles with scalability due to the computational overhead of symbolic execution, limiting its applicability in time-sensitive or large-scale scenarios.

#### 6.3.3 *LXMERT.* 
LXMERT (Tan and Bansal 2019b) utilizes cross-modality transformers to jointly encode visual and textual inputs, facilitating strong multimodal understanding. Although LXMERT was not originally designed for explicit commonsense integration, its language backbone implicitly encodes commonsense priors learned from large text corpora. It achieves high accuracy on VQA and image-text retrieval tasks and is highly scalable owing to its modular transformer architecture, making it suitable for real-world deployment.

#### 6.3.4 *ViLT.* 
ViLT (W. Kim et al. 2021b) further streamlines multimodal learning by removing heavy visual feature extraction stages, directly applying transformers to image patches and text tokens. By leveraging pretrained language models such as BERT (Devlin et al. 2019a), ViLT captures commonsense knowledge without explicit external graphs. It achieves state-of-the-art results on multiple benchmarks. However, training and inference costs remain high, and the model's reliance on dense data makes it less efficient for edge deployment.

#### 6.3.5 *Graph R-CNN.* 
Graph R-CNN (Yang et al. 2018b) leverages scene graphs to explicitly model object-object relationships in images, enhancing relational reasoning capabilities. While this method provides strong commonsense modeling through spatial and semantic relationships, generating and maintaining scene graphs incurs significant computational costs. Consequently, Graph R-CNN exhibits limited scalability and slower inference, reducing its applicability for real-time or highly dynamic environments.

## 7 Recent Trends in Commonsense Reasoning (2022–2025)

In recent years, commonsense reasoning in computer vision has undergone a significant transformation, driven largely by the advent of large multimodal pretraining, retrieval-augmented methods, and foundation models that unify vision, language, and external knowledge. While earlier works relied heavily on static scene graphs or direct knowledge integration, modern models focus on scaling commonsense reasoning through flexible architectures that learn from vast corpora with implicit or explicit retrieval mechanisms. This section offers a comprehensive review of the most significant developments from 2022 to 2024.

### 7.1 Retrieval-Augmented Vision Reasoning

The REVIVE approach (REgional VIsual Representation for knowledge-based Visual question answering) (Y. Lin et al. 2022b) revolutionizes external knowledge retrieval in VQA by emphasizing region-aware visual grounding and dynamic knowledge integration. This approach emphasizes the importance of integrating structured knowledge sources, such as knowledge graphs and Wikidata, to enhance the contextual understanding of visual data and facilitate more accurate predictions in real-world applications. REVIVE performs external knowledge retrieval through a sophisticated, multi-stage process that dynamically integrates structured knowledge bases with visual-language understanding. This approach demonstrates that utilizing regional visual information more effectively significantly improves the performance of knowledge-based VQA models. **Figure 10** demonstrates the key components and workflow of REVIVE.

### 7.2 Foundation Models for Vision-Language Commonsense

OmniVL (Omni-modality Vision-Language model) (X. Wang et al. 2023) extends the idea of foundation models to jointly handle image, video, and text modalities under a single architecture. Trained on billions of multimodal samples, OmniVL demonstrates that large-scale pretraining naturally imbues models with commonsense knowledge by learning visual-language correspondences at scale. Crucially, OmniVL achieves state-of-the-art results across image captioning, visual entailment, and grounded commonsense inference tasks, demonstrating that massive pretraining can serve as an implicit mechanism for commonsense learning. Figure 11 demonstrates the workflow of OmniVL.

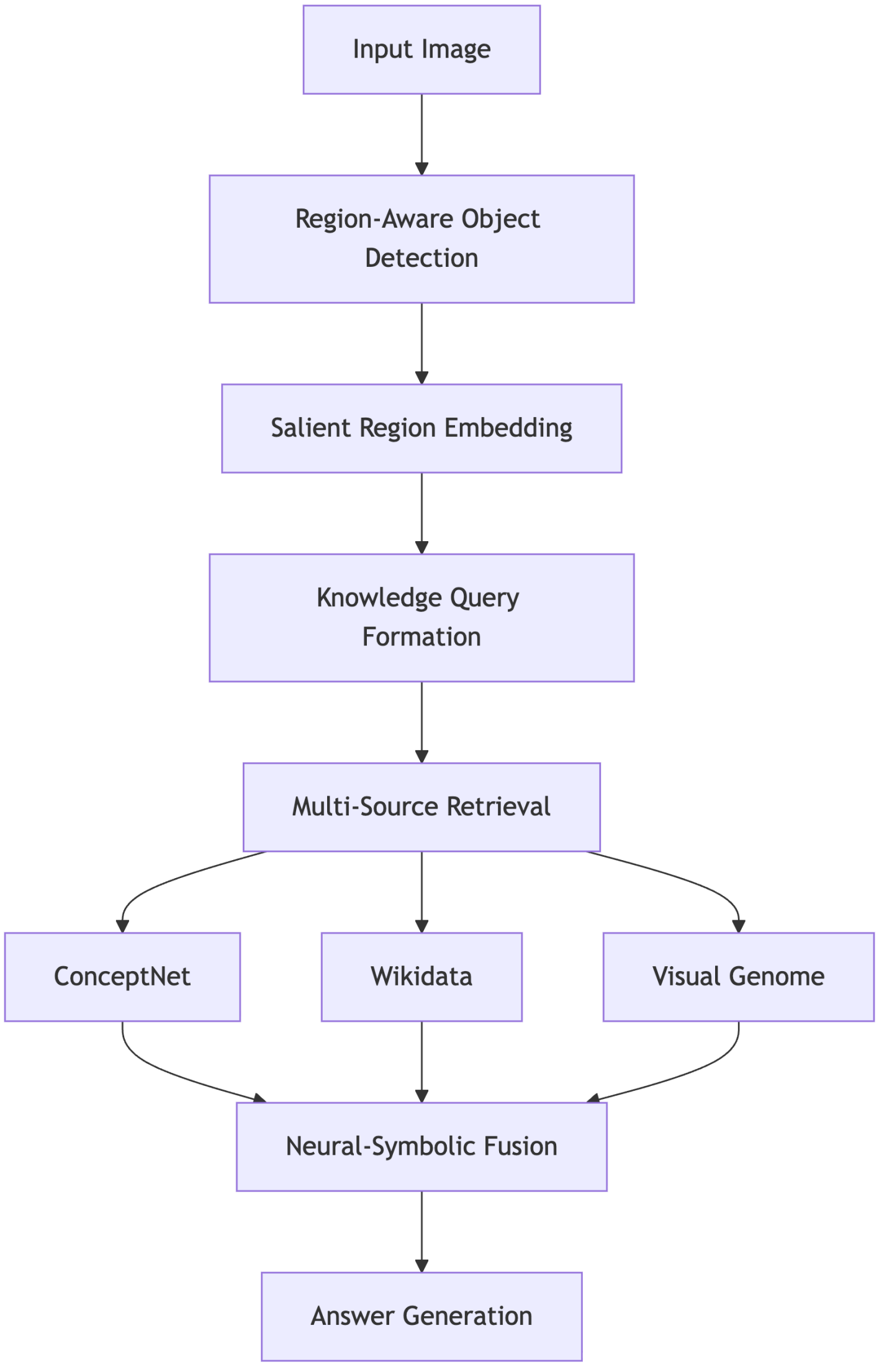

Fig. 10. Graphical representation of the key components and dynamic knowledge retrieval process of the REVIVE framework for knowledge-based visual question answering.

## 7.3 Unified Multimodal Pretraining

The unified multimodal pretraining approach represents a significant advancement in commonsense reasoning, enabling models to effectively learn from diverse data sources and enhance their contextual understanding across various tasks. BEiT-3 (Bidirectional Encoder representation from Image-Text pretraining) (W. Wang et al. 2023) proposes a unified vision-language-text model that extends masked language modeling to images. By treating image patches and text tokens equally, BEiT-3 enables bidirectional context reasoning across modalities, including implicit commonsense relationships. Unlike earlier multimodal models, BEiT-3 integrates visual and textual context at the token level, allowing a deeper, compositional understanding of commonsense. Figure 12 demonstrates the workflow of BEIT.

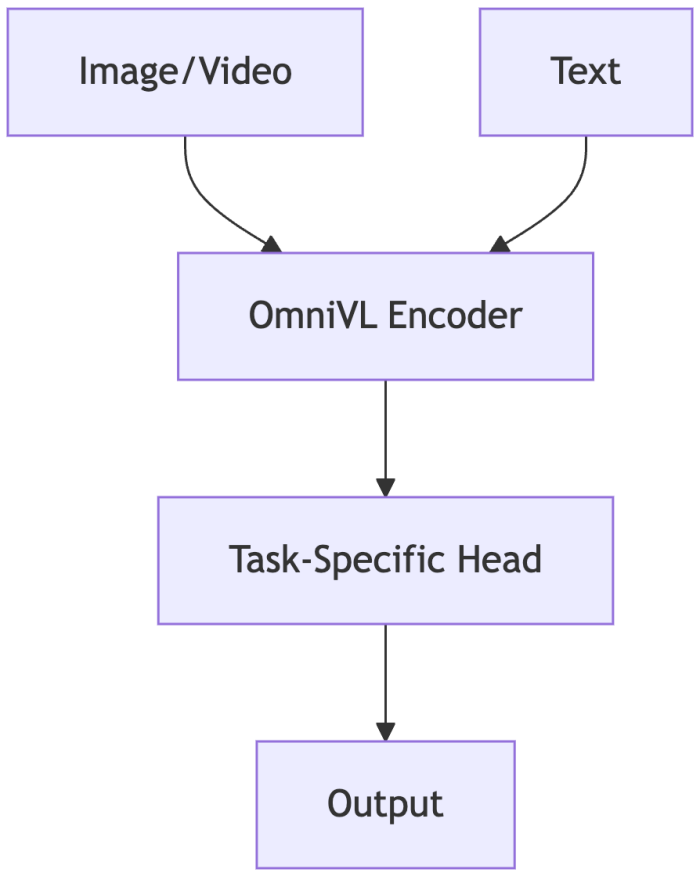

Fig. 11. Graphical representation of the workflow of an omni-modality vision–language model integrating multiple modalities for unified visual and linguistic understanding.

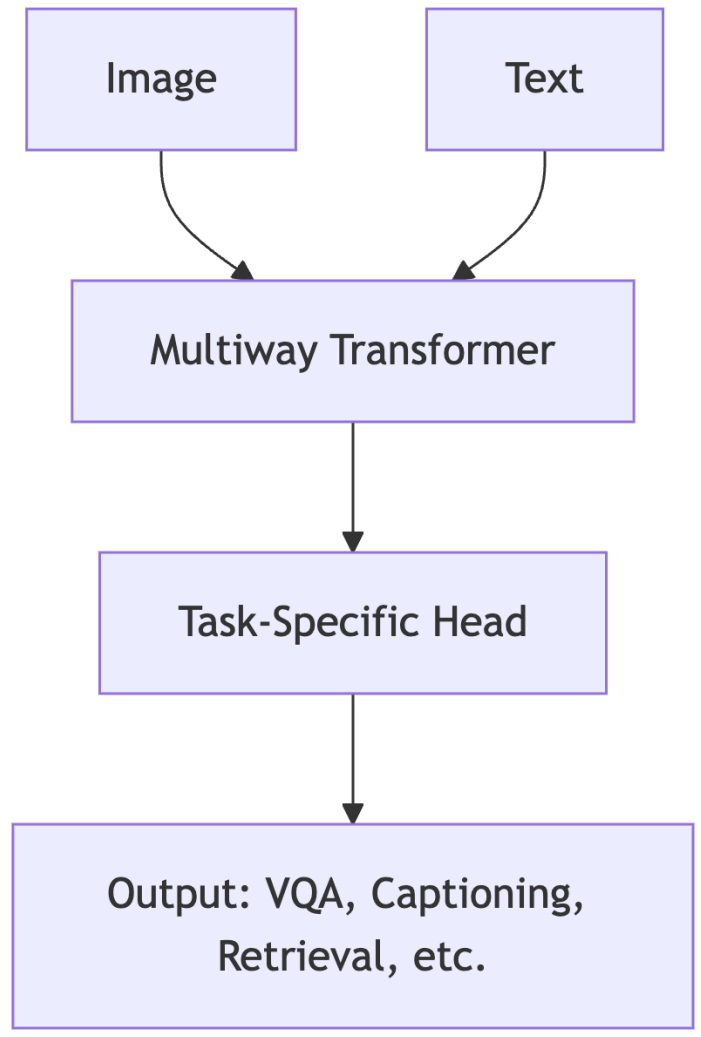

Fig. 12. Graphical representation of the workflow and architecture of BEiT-3 for unified vision–language pretraining.

### 7.4 Object-Semantics Grounded Commonsense

OSCAR++ (**li2022oscarpp**) advances object-centric vision-language pretraining by explicitly grounding linguistic phrases in detected object regions. By aligning object tags with textual input, OSCAR++ strengthens the model's ability to reason about object affordances and relationships, a key aspect of commonsense reasoning. This grounding enables models to understand better interactions like "a knife is used for cutting" even when the action is not visually explicit. Figure 13 demonstrates the workflow of OSCAR++.

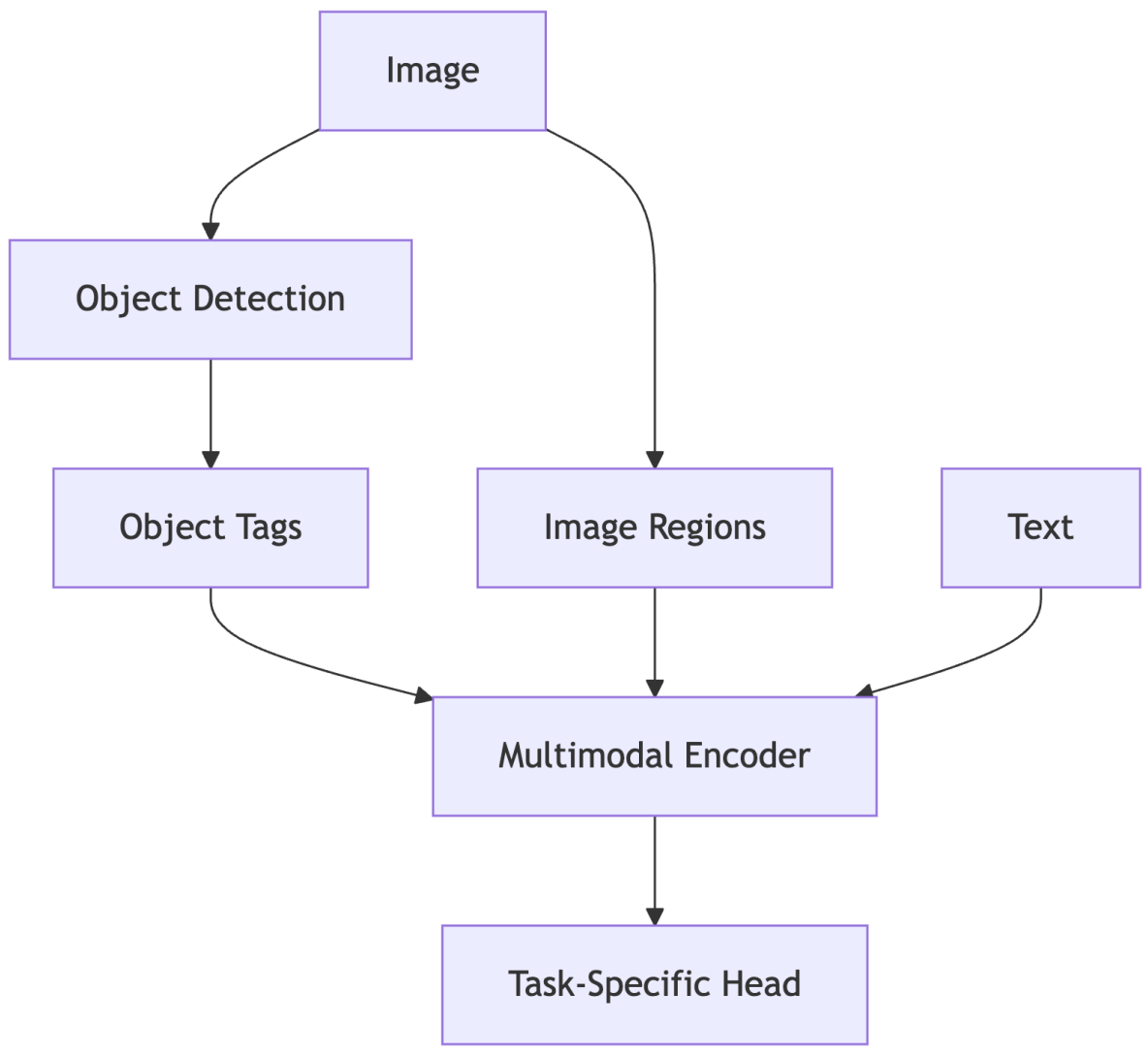

Fig. 13. Graphical representation of the workflow and architecture of OSCAR++ for grounded vision–language pretraining.

### 7.5 Few-Shot Vision-Reasoning Models

Flamingo (Alayrac et al. 2022) introduces a new paradigm for few-shot visual reasoning by conditioning a frozen language model on a small number of vision-language examples. Despite seeing very few examples, Flamingo achieves competitive performance on visual commonsense tasks, suggesting that pretrained language models inherently carry transferable commonsense priors that can be rapidly adapted to new visual domains. The Flamingo represents a shift towards *few-shot commonsense adaptation* in vision models. Figure 14 demonstrates the workflow of Flamingo.

### 7.6 Critical Insights

Compared to earlier methods (2019–2021) that integrated commonsense knowledge either statically (e.g., scene graphs) or symbolically (e.g., neuro-symbolic models), recent trends focus on *dynamic retrieval*, *implicit commonsense learning through scale*, and *object-grounded vision-language alignment*. Retrieval-augmented and foundation models demonstrate that scaling data and model size alone can yield significant commonsense capabilities, although explicit symbolic reasoning remains limited. Moreover, while few-shot methods like Flamingo promise rapid adaptation, they still rely heavily on pretraining biases and have yet to match the robustness of neuro-symbolic approaches in complex logical reasoning.

Despite substantial progress, challenges remain in ensuring factual consistency, interpretability of retrieved knowledge, and reasoning about abstract, non-visual commonsense (e.g., human intentions, physical affordances under novel conditions).

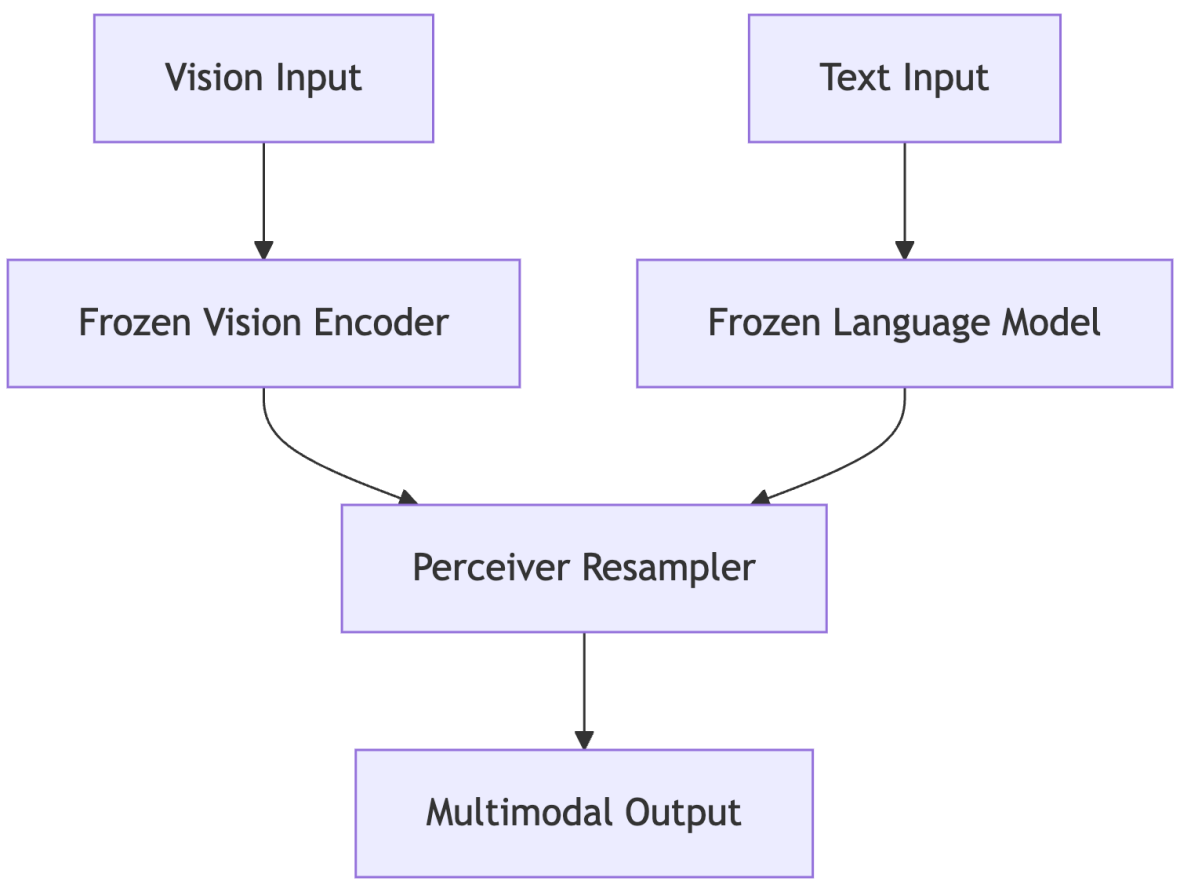

Fig. 14. Graphical representation of the workflow and architecture of the Flamingo vision–language model for few-shot multimodal learning.

## 8 Challenges and Open Problems

Despite recent advancements in commonsense reasoning for computer vision, significant challenges persist that prevent current models from achieving human-like understanding and reasoning capabilities. This section critically discusses the major open problems identified in the recent literature.

### 8.1 Reasoning Beyond Observables

Many current models excel at surface-level reasoning over visual content but struggle to infer unobservable factors such as intentions, goals, emotions, or causality. For instance, even advanced models like OmniVL (X. Wang et al. 2023) primarily rely on observable patterns rather than latent causal structures. Human commonsense reasoning fundamentally depends on inferring hidden states and predicting future outcomes based on partial observations. Enabling models to reason beyond the visible, particularly regarding implicit human activities, remains an unresolved problem that requires advances in causal modeling and abductive inference.

### 8.2 Scalability and Efficiency in Real-Time Systems

While foundation models such as BEiT-3 (W. Wang et al. 2023) and Flamingo (Alayrac et al. 2022) show impressive commonsense capabilities, their enormous size and computational demands hinder deployment in real-world, latency-sensitive applications like autonomous driving or human-robot interaction. Commonsense-enabled reasoning modules must balance the trade-off between *expressive power* and *computational feasibility*. Designing lightweight architectures or retrieval-augmented modules that operate under strict computational budgets without sacrificing reasoning depth remains a critical open problem.

### 8.3 Biases and Hallucinations in Retrieved Commonsense

Knowledge bases and pretrained models inevitably inherit biases from their training corpora. More critically, retrieval-augmented vision–language systems can amplify rather than mitigate these biases. Retrieval from large, web-based corpora further introduces hallucinations—plausible yet factually incorrect information that is treated as authoritative during inference (Ji et al. 2023; S. Lin et al. 2022).

*Concrete Failure Examples:* **Medical AI Hallucination:** In medical visual question answering systems, retrieval-augmented models may access outdated or incorrect clinical knowledge. A documented failure mode involves a system retrieving obsolete treatment guidelines from web corpora and recommending contraindicated interventions (Bender et al. 2021). Unlike human clinicians, who validate recommendations against current standards of care, such systems treat retrieved content as ground truth without temporal or source verification.

**Autonomous Vehicle Misinterpretation:** In autonomous driving, unfamiliar regional road signage may trigger erroneous commonsense retrieval. For example, querying visually similar signs described in online forums can lead a system to misinterpret a “Yield” sign as “Stop,” causing unnecessary emergency braking and increasing rear-end collision risk (Bojarski, Del Testa, et al. 2016). This failure illustrates the lack of source credibility assessment in retrieval-based commonsense reasoning.

**Gender Bias Amplification:** Commonsense knowledge bases such as ConceptNet encode stereotypical associations (e.g., `<doctor, IsA, male>` and `<nurse, IsA, female>`) (Speer et al. 2017b). When retrieval-augmented VQA models rely on such associations, performance degrades on counter-stereotypical cases. Empirical studies report a 15–20% accuracy drop for images depicting female surgeons or male nurses compared to visual-only baselines (J. Zhao et al. 2017). In these cases, commonsense retrieval exacerbates bias rather than correcting it.

*Why Retrieval Amplifies Bias:* Bias amplification arises from several systemic mechanisms:

(1) **Confirmation Bias.** Models preferentially generate retrieval queries that reinforce existing priors, surfacing evidence consistent with learned associations (Nickerson 1998).
(2) **Lack of Source Verification.** Retrieved passages are treated as factual, allowing a single incorrect or outdated source to propagate through many downstream inferences (Petroni et al. 2019).
(3) **Bias Compounding.** Multi-hop reasoning chains multiply biases across retrieval steps, magnifying errors at each stage (Qi et al. 2021).

Table 6. Contrasting bias and hallucination characteristics across commonsense reasoning paradigms.

| Aspect | Symbolic / Retrieval Systems | Foundation Models |
|---|---|---|
| Bias Source | Explicit in knowledge bases; auditable | Implicit in parameters; opaque |
| Hallucination Mode | Retrieves incorrect facts with traceable sources | Generates false information without verifiable provenance |
| Failure Behavior | Fails explicitly when knowledge is missing | Fails silently with high confidence |
| Correction Path | KB editing or source filtering | Retraining or complex prompt engineering |

*Symbolic Precision vs. Foundation Model Brittleness:* The field has traded explicit, correctable bias in symbolic systems for implicit, harder-to-correct bias in foundation models without resolving the underlying issue (Marcus 2020; Pearl and Mackenzie 2018). Both paradigms remain brittle, differing primarily in how failures manifest.

*When Commonsense Priors Degrade Performance:* Commonsense priors can actively harm performance when they conflict with visual evidence:

- **Medical Imaging.** Strong anatomical priors (e.g., “the heart is left-of-center”) cause misclassification of rare variants such as dextrocardia, reducing accuracy by up to 12% compared to baseline models (Oakden-Rayner et al. 2020).

- **Autonomous Driving.** Retrieved priors linking traffic to braking override contradictory visual cues, leading to unnecessary defensive maneuvers (Bojarski, Yeres, et al. 2017).
- **VQA Stereotypes.** Gender stereotypes dominate contextual reasoning, lowering accuracy for counter-stereotypical images when retrieval is enabled (Hendricks et al. 2018; J. Zhao et al. 2017).

### 8.4 Unified Reasoning Across Modalities

Despite the success of vision-language pretraining, existing models still struggle with deeply unified commonsense reasoning across multiple modalities, including images, videos, audio, and text. Models like OmniVL attempt this by pretraining on mixed modalities, yet fully coherent reasoning across dynamic visual scenes and narrative text remains elusive. Building architectures capable of performing temporally aware, multimodal commonsense reasoning—such as understanding a video scene and relating it to unseen textual commonsense—is a significant open challenge that demands new fusion techniques and pre-training paradigms.

### 8.5 Interpretability and Explainability

As models integrate more external knowledge and perform complex reasoning steps, their decision-making becomes increasingly opaque. The black-box nature of deep commonsense reasoning raises concerns in domains that require transparency, such as healthcare, education, and law. Few current models offer interpretable reasoning chains or traceable commonsense application. Developing commonsense-enabled vision models with built-in explainability, where users can inspect retrieved knowledge, reasoning steps, and final predictions, is critical for trust, accountability, and broader societal acceptance.

### 8.6 Failure Cascades and Error Propagation

Commonsense-enhanced vision systems are vulnerable to *failure cascades*, in which errors introduced during perception propagate and amplify through retrieval and reasoning stages (Amodei et al. 2016; H. Jiang et al. 2020). In most current pipelines, each module assumes the correctness of upstream outputs, leading to compounding failures rather than corrective reasoning.

*Cascade Example: Medical Diagnosis:* A representative cascade occurs in medical decision support systems. A segmentation model fails to detect a small lung nodule due to low contrast. The retrieval module then queries for "small lung nodules" and retrieves population-level commonsense suggesting benignity (Ardila et al. 2019). The reasoning stage amplifies this assumption and concludes that the finding is not clinically concerning. As a result, the system recommends no follow-up, missing an early-stage cancer. The root cause is the absence of uncertainty propagation across stages.

*Cascade Example: Autonomous Driving:* In autonomous driving, an object detector may misclassify a plastic bag as a small animal. Commonsense retrieval associates animals on the roadway with emergency braking behavior. The planning module executes a hard braking maneuver, increasing rear-end collision risk (Bojarski, Del Testa, et al. 2016; Bojarski, Yeres, et al. 2017). Incorrect commonsense is amplified rather than corrected, despite weak or ambiguous perceptual evidence.

*Why Failure Cascades Occur:* Failure cascades persist because current systems lack key safeguards:

- **No uncertainty propagation**: Confidence estimates are not transferred between perception, retrieval, and reasoning modules (Gal and Ghahramani 2016; Guo et al. 2017).
- **No conflict detection**: Systems cannot identify contradictions between retrieved commonsense and visual evidence.
- **No backtracking**: Once a hypothesis is selected, models cannot revise earlier decisions.

Table 7. Failure modes in commonsense-enhanced vision systems:

| Failure Type | Definition | Example | Needed Solution |
|---|---|---|---|
| Prior–Evidence Conflict | Commonsense contradicts perception | Red sunset vs. “sky is blue” prior | Conflict detection; evidence weighting |
| Incomplete Knowledge | Missing relevant facts | Rare disease or cultural practice | Explicit uncertainty; abstention |
| Outdated Knowledge | Temporally invalid commonsense | Pre-pandemic norms on mask usage (Luccioni and Viviano 2021) | Temporal grounding; knowledge versioning |
| Cascading Errors | Early error propagates | Misdetection → wrong retrieval → bad inference | Uncertainty propagation; error detection |

*Taxonomy of Failure Modes:* Current systems are optimized for average-case performance, where commonsense priors align with perceptual evidence, but fail catastrophically on edge cases where they diverge (Hendrycks and Dietterich 2019). This brittleness makes existing commonsense reasoning pipelines unsuitable for safety-critical applications, where tail-risk performance is paramount.

## 9 Future Directions

As commonsense reasoning in computer vision continues to evolve, several promising research avenues are emerging that could bridge the remaining gaps between current models and truly human-level reasoning. Building on the strengths and challenges outlined in recent work, this section highlights key directions for future research.

### 9.1 Commonsense-Infused Causal Modeling

Reasoning beyond direct visual observations demands a deeper integration of causal commonsense knowledge. Current models primarily rely on statistical correlations, but human commonsense fundamentally involves predicting unobserved causes and effects. Future work should focus on embedding *causal graphs*, *counterfactual reasoning*, and *abductive inference* into vision-language models. Integrating causality-driven frameworks into pretraining or fine-tuning could enable models to infer latent intentions, action consequences, and goal-directed behaviors from visual scenes.

### 9.2 Scalable Lightweight Reasoning Modules

With the increasing demand for commonsense reasoning in real-time, embedded, and mobile applications, there is an urgent need to develop lightweight commonsense reasoning modules. Future research should explore *knowledge distillation* of large commonsense models into compact student networks, *on-device retrieval augmentation*, and *low-rank adaptation techniques* for foundation models. Balancing the richness of reasoning with computational efficiency will be crucial for deploying commonsense-enabled vision models in safety-critical domains, such as autonomous vehicles, robotics, and augmented reality.

### 9.3 Bias Mitigation and Trustworthy Commonsense Integration

Commonsense knowledge bases and large-scale pretraining data often contain societal biases and inaccuracies. Future commonsense-enhanced vision systems must incorporate mechanisms for *bias detection*, *knowledge verification*, and *trust calibration*. Techniques such as adversarial robustness training, factual consistency regularization, and ethical retrieval filtering will be crucial in ensuring that retrieved knowledge supports fair, safe, and unbiased decision-making. Additionally, interpretability tools that expose which knowledge pieces influenced a model's prediction will be essential for trust and transparency.

### 9.4 Unified Multimodal Commonsense Reasoning

The next frontier involves achieving coherent commonsense reasoning across diverse modalities, including vision, language, video, audio, and 3D spatial understanding. Early foundation models, such as OmniVL (X. Wang et al. 2023) and Flamingo (Alayrac et al. 2022), demonstrate the potential of large-scale multimodal pretraining; however, fully unified commonsense reasoning across time, space, and modalities remains elusive. Future research must address how to build models capable of dynamically grounding and reasoning about commonsense concepts across complex, multisensory inputs, enabling applications such as embodied AI, virtual assistants, and interactive storytelling.

### 9.5 Explainable Commonsense Reasoning

Finally, explainability will become a first-class objective in commonsense reasoning models. Future architectures should generate *interpretable reasoning traces* that show the retrieved knowledge, intermediate inferences, and causal chains leading to a prediction. Techniques such as *commonsense chain-of-thought prompting*, *retrieval visualization*, and *symbolic post-hoc explanation* will be crucial in making commonsense-enabled models more transparent, accountable, and suitable for deployment in critical settings, including healthcare, education, and law.

## 10 Conclusion

Commonsense reasoning represents a critical frontier in advancing the capabilities of computer vision systems beyond pattern recognition toward accurate contextual understanding. In this survey, we systematically analyzed the evolution of commonsense integration in vision, from early scene graphs and neuro-symbolic models to modern retrieval-augmented and foundation model approaches. Through detailed comparative analysis, we highlighted how different paradigms address the challenges of scalability, reasoning depth, and real-world applicability.

Recent innovations between 2022 and 2025, such as retrieval-augmented commonsense reasoning (REVIVE), unified multimodal pretraining (BEiT-3, OmniVL), and few-shot adaptation (Flamingo), signal a paradigm shift toward flexible, dynamic reasoning architectures capable of leveraging external knowledge in real time. However, significant open challenges remain, including the need for dynamic knowledge retrieval, causal inference beyond observables, computational scalability, bias mitigation, multimodal unification, and transparent explainability.

Future research must bridge the strengths of symbolic reasoning, retrieval-augmented architectures, and foundation models to build vision systems that reason like humans do, flexibly, contextually, and ethically. Dynamic commonsense retrieval, lightweight causal reasoning modules, bias-aware learning, and interpretable decision-making are essential to achieve trustworthy, robust, and generalizable commonsense reasoning.

As the field advances, commonsense-enabled vision will become foundational not only for classical tasks such as recognition and captioning but also for emerging domains like autonomous agents, embodied AI, assistive robotics, and human-AI collaboration. By addressing the current limitations and embracing interdisciplinary approaches, the vision community stands poised to realize the full promise of human-like commonsense reasoning in artificial intelligence.

References

-e

## A Reproducibility Checklist

Select the answers that apply to your research – one per item.

### All articles:

(1) All claims investigated in this work are clearly stated. [yes]
(2) Clear explanations are given how the work reported substantiates the claims. [yes]
(3) Limitations or technical assumptions are stated clearly and explicitly. [yes]
(4) Conceptual outlines and/or pseudo-code descriptions of the AI methods introduced in this work are provided, and important implementation details are discussed. [NA]
(5) Motivation is provided for all design choices, including algorithms, implementation choices, parameters, data sets and experimental protocols beyond metrics. [NA]

### Articles containing theoretical contributions:

Does this paper make theoretical contributions? [no]

### Articles reporting on computational experiments:

Does this paper include computational experiments? [no]

### Articles using data sets:

Does this work rely on one or more data sets (possibly obtained from a benchmark generator or similar software artifact)? [yes]

If yes, please complete the list below.

(1) All newly introduced data sets are included in an online appendix or will be made publicly available upon publication of the paper. The online appendix follows best practices for long-term accessibility with a license that allows free usage for research purposes. [NA]
(2) The newly introduced data set comes with a license that allows free usage for reproducibility purposes. [NA]
(3) The newly introduced data set comes with a license that allows free usage for research purposes in general. [NA]
(4) All data sets drawn from the literature or other public sources (potentially including authors' own previously published work) are accompanied by appropriate citations. [yes]
(5) All data sets drawn from the existing literature (potentially including authors' own previously published work) are publicly available. [yes]
(6) All new data sets and data sets that are not publicly available are described in detail, including relevant statistics, the data collection process and annotation process if relevant. [NA]
(7) All methods used for preprocessing, augmenting, batching or splitting data sets (e.g., in the context of hold-out or cross-validation) are described in detail. [NA]

### Explanations on any of the answers above (optional):

This is a comprehensive survey paper on commonsense reasoning in computer vision. The paper does not introduce new theoretical contributions, computational experiments, or novel data sets. Instead, it provides a systematic review and comparative analysis of existing approaches, models, data sets, and evaluation benchmarks from the literature. All cited data sets (e.g., Visual Commonsense Reasoning, WHOOPS!, SIEVE-Bench) are publicly available and appropriately referenced. The paper's contributions lie in its analytical synthesis, taxonomy of approaches, and identification of open challenges in the field.